\documentclass[10pt,twocolumn,letterpaper]{article}

\usepackage{cvpr}
\usepackage{times}
\usepackage{epsfig}
\usepackage{graphicx}
\usepackage{amsmath}
\usepackage{amssymb}

\usepackage[pagebackref=true,breaklinks=true,letterpaper=true,colorlinks,bookmarks=false]{hyperref}

\cvprfinalcopy
\begin{document}
	
	\title{PointASNL: Robust Point Clouds Processing using Nonlocal Neural Networks with Adaptive Sampling}
	
	
	\author{Xu Yan$^{1,2}$\quad Chaoda Zheng$^{2,3}$\quad Zhen Li$^{1,2,}$\thanks{{ Corresponding author: Zhen Li.}}\quad Sheng Wang$^4$\quad Shuguang Cui$^{1,2}$ \\
		$^1$ The Chinese University of Hong Kong (Shenzhen), $^2$Shenzhen Research Institute of Big Data \\ 
		$^3$South China University of Technology, $^4$Tencent AI Lab\\
	{\tt\small	\{xuyan1@link., \textbf{lizhen@}, shuguangcui@\}cuhk.edu.cn}}
	\maketitle
	
	\begin{abstract}

		Raw point clouds data inevitably contains outliers or noise through acquisition from 3D sensors or reconstruction algorithms. In this paper, we present a novel end-to-end network for robust point clouds processing, named PointASNL, which can deal with point clouds with noise effectively. The key component in our approach is the adaptive sampling (AS) module. It first re-weights the neighbors around the initial sampled points from farthest point sampling (FPS), and then adaptively adjusts the sampled points beyond the entire point cloud. Our AS module can not only benefit the feature learning of point clouds, but also ease the biased effect of outliers. To further capture the neighbor and long-range dependencies of the sampled point, we proposed a local-nonlocal (L-NL) module inspired by the nonlocal operation. Such L-NL module enables the learning process insensitive to noise. Extensive experiments verify the robustness and superiority of our approach in point clouds processing tasks regardless of synthesis data, indoor data, and outdoor data with or without noise. Specifically, PointASNL achieves state-of-the-art robust performance for classification and segmentation tasks on all datasets, and significantly outperforms previous methods on real-world outdoor SemanticKITTI dataset with considerate noise. Our code is released through {\url{https://github.com/yanx27/PointASNL}}.
		
		
		
	\end{abstract}
	
	\vspace{-0.3cm}
	\section{Introduction}
	
	
	With the popularity of 3D sensors, it's relatively easy for us to obtain more raw 3D data, e.g., RGB-D data, LiDAR data, and MEMS data~\cite{PointConv}. Considering point clouds as the fundamental representative of 3D data, the understanding of point clouds has attracted extensive attention for various applications, e.g., autonomous driving~\cite{shi2019pointrcnn}, robotics~\cite{wang2015voting}, and place recognition~\cite{liu2019lpd}.  Here, a point cloud has two components: the points $\mathcal{P} \in \mathbb{R}^{N\times	3}$ and the features $\mathcal{F} \in \mathbb{R}^{N\times	D}$. Unlike 2D images, the sparsity and disorder proprieties make robust point clouds processing a challenging task. Furthermore, the raw data obtained from those 3D sensors or reconstruction algorithms inevitably contain outliers or noise in real-world situations.

	
	\begin{figure}[t]
		\begin{center}
			\includegraphics[width=\columnwidth]{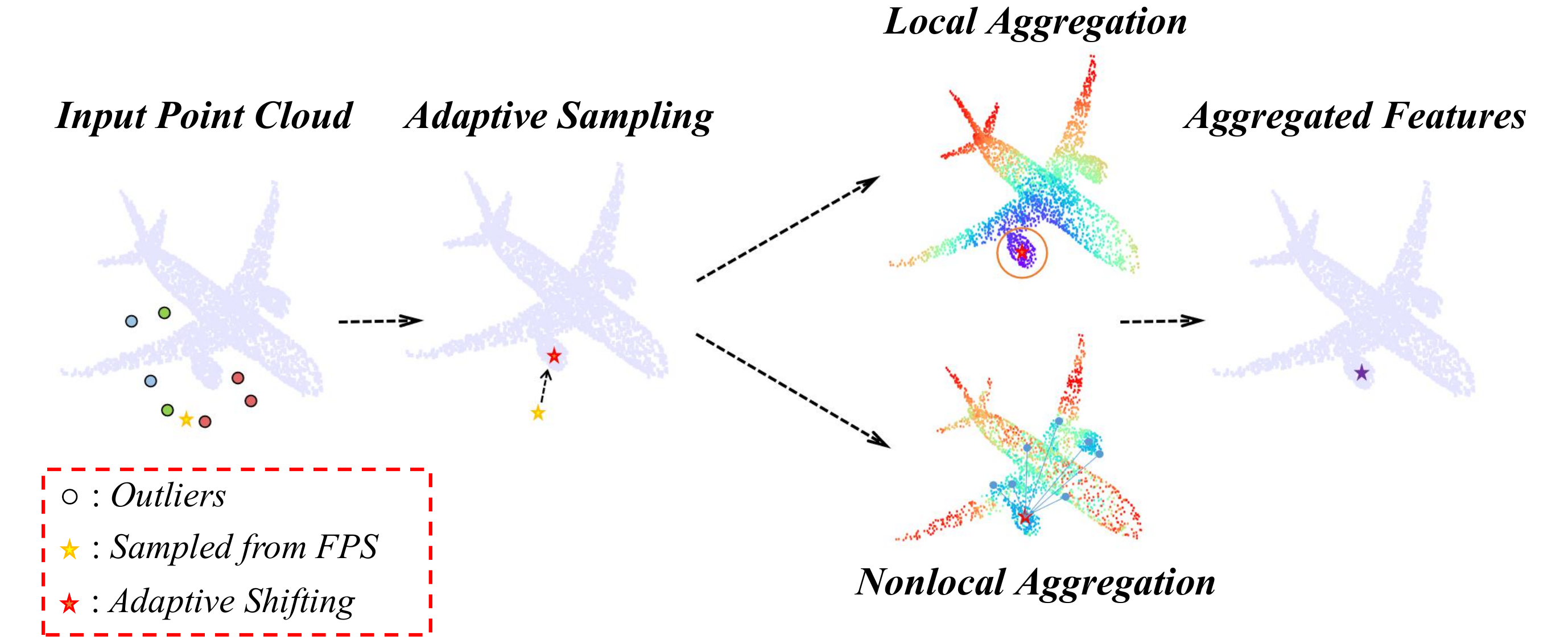}
		\end{center}
		\caption{{\bf PointASNL for robust point clouds processing}. The adaptive sampling module adaptively adjusts the sampled point from point clouds with noise. Besides, the local-nonlocal module not only combines the local features in Euclidean space, but also considers the long-range dependency in feature space.}
		\label{fig:fig1}
		\vspace{-0.5cm}
	\end{figure}
	
	In this work, we present a novel end-to-end network for robust point clouds processing, named PointASNL, which can deal with point clouds with noise or outliers effectively. Our proposed PointASNL mainly consists of two general modules: adaptive sampling (AS) module and local-nonlocal (L-NL) module. The AS module is used to adjust the coordinates and features of the sampled points, whereas the L-NL module is used to capture the neighbor and long-range dependencies of the sampled points.
	
	Unlike the cases in 2D images, traditional convolution operations cannot directly work on unstructured point cloud data. Thus, most of the current methods usually use sampling approaches to select points from the original point clouds for conducting local feature learning. Among these sampling algorithms, farthest point sampling (FPS)~\cite{pointnet2}, Poisson disk sampling (PDS)~\cite{Monte}, and Gumbel subset sampling (GSS)~\cite{Gumbel} are proposed in previous works. However, as the most representative one, FPS is rooted in Euclidean distance, which is task-dependent and outliers sensitive. PDS, a predefined uniformly sampling method, also cannot solve the problem above in a data-driven way. GSS only performs sampling from a high-dimension embedding space and ignores the spatial distribution of points. Furthermore, the shared key issue in these approaches is that the sampled points are limited to a subset of the original point clouds. Therefore, as shown in the left part of Fig.~\ref{fig:fig1}, suppose an outlier point is sampled, it will influence the downstream process inevitably.
	
	To overcome the issues mentioned above, we propose a differentiable adaptive sampling (AS) module to adjust the coordinates of the initial sampled points (e.g., from FPS) via a data-driven way. Such coordinate adjusting facilitates to fit the intrinsic geometry submanifold and further shifts to correct points beyond original point clouds without the influence of outliers. Thus, the AS module can not only benefit point feature learning, but also improve the model robustness to noise.

	To further enhance the performance as well enables the learning process insensitive to noise, we proposed a local-nonlocal (L-NL) module for capturing neighbor and long-range dependencies of the sampled points. The underlying reason is that, currently, most appealing methods for feature learning is to query a local group around the each sampled point, and then they construct the graph-based learning~\cite{ecc,DGCNN,PointWeb, HPEIN} or define convolution-like operations~\cite{Pointwise,Spidercnn,groh2018flex,PCNN,PointConv,KPCONV} (we denote them as \textit{Point Local Cell}). Nonetheless, such point local cell only considers local information interaction in the neighbor area and then acquires the global context through a hierarchical structure, which usually leads to bottom-up feature learning. Inspired by the success of the Nonlocal network~\cite{Nonlocal}, we innovatively design this L-NL module, in which the key component is the \textit{Point Nonlocal Cell}. In particular, the point nonlocal cell allows the computation of the response of a sampled point as a weighted sum of the influences of the entire point clouds, instead of just within a limited neighbor range. With the learned long-dependency correlation, the L-NL module can provide more precise information for robust point clouds processing. As shown in the right part of Fig.~\ref{fig:fig1}, although the sampled points within the lower engine are covered with noise, our L-NL module can still learn the features from the other engine with a different noise distribution.
	
	
	
	Our main contribution can be summarized as follows: 1) We propose an end-to-end model for robust point clouds processing, PointASNL, which can effectively ease the influence of outliers or noise; 2) With the proposed adaptive sampling (AS) module, PointASNL can adaptively adjust the coordinates of the initial sampled points, making them more suitable for feature learning with intrinsic geometry and more robust for noisy outliers; and 3) We further design a point nonlocal cell in the proposed local-nonlocal (L-NL) module, which enhances the feature learning in point local cells. Extensive experiments on classification and segmentation tasks verify the robustness of our approach.


	
	\section{Related Work}
	

	\begin{figure*}
		\begin{center}
			\begin{center}
				\includegraphics[width=0.95\linewidth]{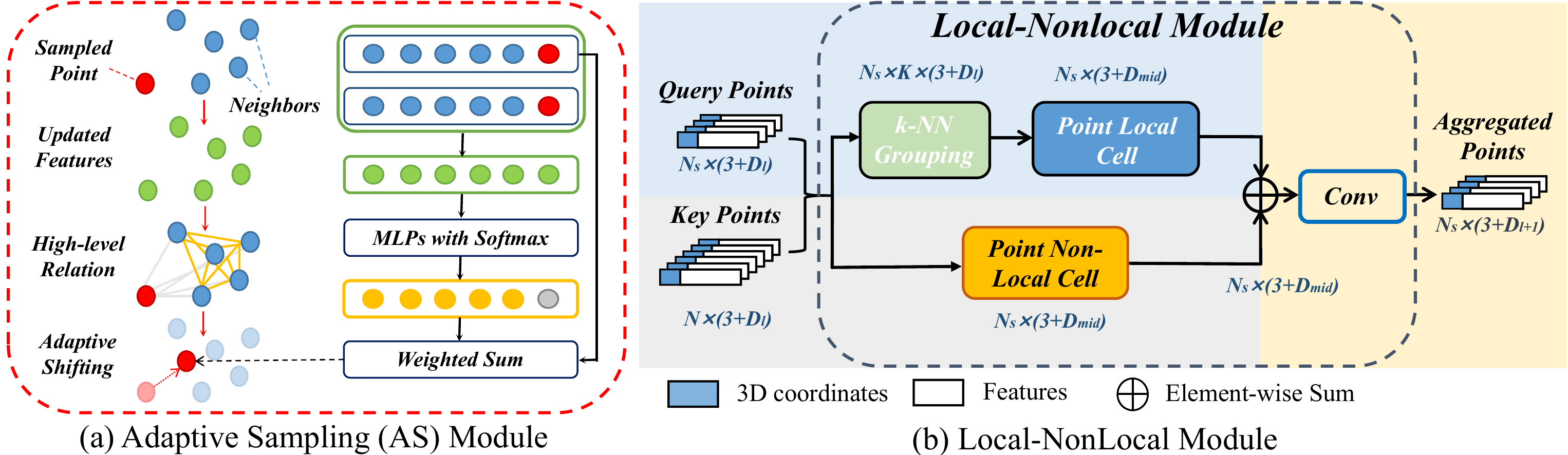}
			\end{center}
		\end{center}
		\caption{Part (a) shows adaptive sampling (AS) module, which firstly updates features of grouping point by reasoning group relationship, then normalized weighs re-weight initial sampled points to achieve new sampled points. Part (b) illustrates the construction of local-nonlocal (L-NL) module, which consists of point local cell and point nonlocal cell. $N_s$ stands for sampled point number, $N$ stands for point number of entire point clouds, $D_l, D_{mid}$, and $D_{l+1}$ stand for channel numbers.}
		\label{fig:ASModule}
		\vspace{-0.3cm}
	\end{figure*}
	{\noindent\bf Volumetric-based and Projection-based Methods.}
	Considering the sparsity of point clouds and memory consumption, it is not very effective to directly voxelized point clouds and then use 3D convolution for feature learning. Various subsequent improvement methods have been proposed, e.g., efficient spatial-temporal convolution MinkowskiNet~\cite{choy20194d}, computational effective Submanifold sparse convolution~\cite{SparseConv}, and Oc-tree based neural networks O-CNN~\cite{wang2017cnn} and OctNet~\cite{riegler2017octnet}. Such methods greatly improve the computational efficiency, thus leading to the entire point clouds as input without sampling and superior capacity. There are also other grid-based methods using traditional convolution operations, e.g., projecting 3D data to multi-view 2D images~\cite{su2015multi} and lattice space~\cite{su2018splatnet}. Yet, the convolution operation of these methods lacks the ability to capture nonlocally geometric features. 
	
	{\noindent\bf Point-based Learning Methods.}
	PointNet~\cite{pointnet} is the pioneering work directly on sparse and unstructured point clouds, which summarizes global information by using pointwise multi-layer perceptions (MLPs) followed by the max-pooling operation.  PointNet++~\cite{pointnet2} further applies a hierarchical structure with k-NN grouping followed by max-pooling to capture regional information. Since it aggregates local features simply to the largest activation, regional information is not yet fully utilized. Recently, much effort has been made for effective local feature aggregation. PointCNN~\cite{PointCNN} transforms neighboring points to the canonical order, which enables traditional convolution to play a normal role. Point2Sequence~\cite{Point2Sequence} uses the attention mechanism to aggregate the information of different local regions. Methods~\cite{Spidercnn,PointConv, Monte,rscnn,ecc, PCCN} directly use the relationship between neighborhoods and local centers to learn a dynamic weight for convolution, where ECC~\cite{ecc} and RS-CNN~\cite{rscnn} use ad-hoc defined 6-D and 10-D vectors as edge relationship, PCCN~\cite{PCCN} and PointConv~\cite{PointConv} project the relative position of two points to a convolution weight. A-CNN~\cite{acnn} uses ring convolution to encode features that have different distances from the local center points, and PointWeb~\cite{PointWeb} further connects every point pairs in a local region to obtain more representative region features. Still, these methods only focus on local feature aggregation and acquire global context from local features through a hierarchical structure. On the other hand, there are various works for learning the global context from the local features. A-SCN~\cite{ASCN} uses a global attention mechanism to aggregate global features but lacks the support of local information, which does not achieve good results. DGCNN~\cite{DGCNN} proposes the EdgeConv module to generate edge features and search neighbors in features space. LPD-Net~\cite{liu2019lpd} further extends DGCNN on both spatial neighbors and features neighbors aggregation. Nonetheless, the neighbors in the feature space are not representative of the global features, and spatial receptive fields of the network gradually become confused without a hierarchical structure.

	{\noindent\bf Outlier Removal and Sampling Strategy.} Outliers and noise usually exist in raw point clouds data. Previous robust statistics methods~\cite{aggarwal2015outlier} for outlier removal suffer from non-trivial parameter tuning or require additional information~\cite{wolff2016point}. Various data-driven methods~\cite{guerrero2018pcpnet,rakotosaona2019pointcleannet} are proposed for outlier removal, which first discard some outliers and then projects noisy points to clean surfaces. Yet, such methods cannot inherently merge the robust point cloud feature learning with outlier removal in a joint learning manner. On the other hand, deep learning based point cloud processing methods usually sample points to decrease computational consumption. Though, most sampling methods are limited by noise sensitivity and not driven by data~\cite{pointnet2,Monte}, or without the consideration of spatial distribution ~\cite{Gumbel}. SO-Net~\cite{So-net} uses an unsupervised neural network, say self-organizing map (SOM), to utilize spatial distribution of point clouds. It then employs PointNet++~\cite{pointnet2} to multiple smaller sampled 'nodes'. However, SO-Net does not belong to online adaptive sampling. Under the assumption of local label consistency, some works use the geometric centers of voxel grids to uniformly represent sampled points~\cite{KPCONV,ecc}, which ignores the difference of point distribution influence. Still, these methods are extremely sensitive to noise and cannot learn the spatial distribution of sampled points at the same time.
	
	

	
	\section{Our Method}
	In this paper, we propose two modules in PointASNL, namely adaptive sampling (AS) module in Sec.~\ref{AS} and local-nonlocal (L-NL) module in Sec.~\ref{NL}. In Sec.\ref{PNL}, we combine AS and L-NL modules in a hierarchical manner to form our proposed PointASNL model.

	\subsection{Adaptive Sampling (AS) Module}
	\label{AS}
	Farthest point sampling (FPS) is widely used in many point cloud framework, as it can generate a relatively uniform sampled points. Therefore, their neighbors can cover all input point clouds as much as possible. Nevertheless, there are two main issues in FPS: (1) It is very sensitive to the outlier points, making it highly unstable for dealing with real-world point clouds data. (2) Sampled points from FPS must be a subset of original point clouds, which makes it challenging to infer the original geometric information if occlusion and missing errors occur during acquisition.

	To overcome the above-mentioned issues, we first use FPS to gain the relatively uniform points as original sampled points. Then our proposed AS module adaptively learns shifts for each sampled point. Compared with the similar process widely used in mesh generation~\cite{wang2018pixel2mesh}, the downsampling operation must be taken into account both in spatial and feature space when the number of points is reduced.  For the AS module, let $\mathcal{P}_{s} \in \mathbb{R}^{N_s\times 3}$ as the sampled $N_s$ points from  $N$ input points of certain layer, $x_i$ from $\mathcal{P}_{s}$ and $f_i$ from $\mathcal{F}_s \in \mathbb{R}^{N_s\times D_l}$ as  a sampled point and its features. We first search neighbors of sampled points as groups via k-NN query, then use general self-attention mechanism~\cite{vaswani2017attention} for group features updating.

	As shown in Fig.~\ref{fig:ASModule}~(a), we update group features by using attention within all group members. For $x_{i,1},..., x_{i,K} \in \mathcal{N}(x_i)$ and their corresponding features $f_{i,1},..., f_{i,K}$, where $\mathcal{N}(x_i)$ is $K$ nearest neighbors of sampled point $x_i$, feature updating of group member $x_{i,k}$ can be written as 
	\begin{equation}
	f_{i,k} =  \mathcal{A}(\mathcal{R}(x_{i,k} ,x_{i,j})\gamma(x_{i,j}),~~ \forall x_{i,j} \in \mathcal{N}(x_i)),
	\label{eqattention}
	\end{equation}
	where a pairwise function $\mathcal{R}$ computes a high level relationship between group members $x_{i,k}, x_{i,j} \in \mathcal{N}(x_i)$. The unary function $\gamma$ change the each group feature $f_{i,j}$ from dimension $D_l$ to another hidden dimension $D'$ and $\mathcal{A}$ is a aggregation function. 
	
	For less computation, we consider $\gamma$ in the form of a linear transformation of point features  $\gamma(x_{i,j}) =W_{\gamma} f_{i,j} $, and relationship function $\mathcal{R}$ is dot-product similarity of two points as follows, 
	\begin{equation}
	\mathcal{R}(x_{i,k} ,x_{i,j}) = {\text{Softmax}}(\phi(f_{i,k})^T \theta(f_{i,j}) /\sqrt{D'}),
	\end{equation}
	where $\phi$ and $\theta$ are independent two linear transformations and can be easily implemented by independent 1D convolution ~~$Conv:\mathbb{R}^{D_{l}} \mapsto \mathbb{R}^{D'} $, where $D_l$ and $D^{'}$ are input and output channel, respectively.

	
	After that, point-wise MLPs, i.e., $\sigma_p$ and $\sigma_f$ with softmax activation function on $K$ group members are used to obtain the corresponding intensity of each point in a group, which can be represented as normalized weights for each coordinate axis and features channel.
	\begin{equation}
	\begin{split}
	&{F_p} = \{\sigma_{p}(f_{i,k})\}_{k=1}^K, ~~~{W}_p =  \text{Softmax}({F_p}),\\
	&{F_f} = \{\sigma_{f}(f_{i,k})\}_{k=1}^K, ~~~{W}_f =  \text{Softmax}({F_f}),
	\end{split}
	\end{equation}
	where ${F_p}, {F_f}, {W}_p, {W}_f \in \mathbb{R}^{K\times 1}$ are outputs of point-wise MLPs and normalized weights after softmax function. Finally, a adaptive shifting on both $K$ neighbors' coordinates from ${X}\in \mathbb{R}^{K\times 3}$ and their features from ${F}\in \mathbb{R}^{K\times D'}$ are implemented by the weighted sum operation. We obtain a new coordinate of the sampled point $x^*_i$ and its features $f^*_i$ by following operations,
	\begin{equation}
	\begin{split}
	&x^*_i =  {W}_p^T {X}, ~~~{X}=\{x_{i,k}\}_{k=1}^K,\\
	&f^*_i =  {W}_f^T {F}, ~~~{F}=\{f_{i,k}\}_{k=1}^K.
	\end{split}
	\end{equation}

	\subsection{Local-Nonlocal (L-NL) Module}
	\label{NL}
	Within our L-NL module, there are two cells: point local (PL) cell and point nonlocal (PNL) cell. Specifically, the PL cell can be any appealing algorithms (e.g., PointNet++~\cite{pointnet2}, PointConv~\cite{PointConv}), and the PNL cell innovatively considers the correlations between sampled points and the entire point cloud in multi-scale. Consequently, the contextual learning of the point cloud is enhanced by combining the local and global information (See Fig.~\ref{fig:ASModule}(b)).
	
	\subsubsection{Point Local Cell} 
	
	The local features mining of point clouds often exploits the local-to-global strategy~\cite{pointnet2}, which aggregates local features in each group and gradually increases the receptive field by hierarchical architectures. We adopt such methods in point local (PL) cell. Similar to the previous definition for a local sampled point $x_i$, corresponding feature $f_i$ and neighborhoods $\mathcal{N}(x_i)$, a generalized local aggregation function used in PL can be formulated as
	
	\begin{equation}
	f_{i}^l =  \mathcal{A}(\mathcal{L}({f_n}), \forall x_n \in \mathcal{N}(x_i)),
	\label{eq1}
	\end{equation}
	where $f^{l}_{i}$ is updated features of local center $x_i$, which is updated by local feature transformation function $\mathcal{L}$ and aggregation function $\mathcal{A}$. For PointNet++~\cite{pointnet2}, $\mathcal{L}$ is multi-layer perceptions (MLPs) and $\mathcal{A}$ is max-pooling. Recently, more and more works directly design convolution operators on the local regions, which mainly change $\mathcal{L}$ to be a learnable weighted multiply obtained by neighbor relationships. Considering the efficiency and effectiveness of the operation in a compromise, we implement the convolution operation by adaptively projecting the relative position of two points to a convolution weight~\cite{PCCN,PointConv}, and aggregate local features,
	\begin{equation}
	\mathcal{L}(f_n) := g(x_n-x_i) f_n,
	\end{equation}
	where $g$ is chosen as MLPs: $\mathbb{R}^3 \mapsto \mathbb{R}^{D_{l}\times D_{mid}}$, which transfers 3 dimension relative position to $D_{l}\times D_{mid}$ transformation matrix. $D_{l}$ represents the channel of the input features in certain layer and $D_{mid}$ is the channel of the updated features by PL cell.
	
	\begin{figure}
		\begin{center}
			\includegraphics[width=0.86 \columnwidth, height=5.2cm]{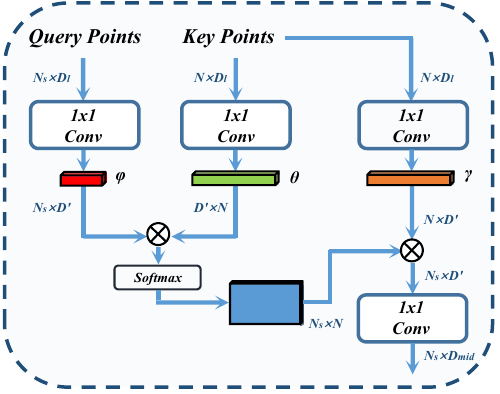}
		\end{center}
		\caption{\textbf{Inner structure of point nonlocal (PNL) cell.} For the notations $N_s, N, D_l, D_{mid}$ please refer to the caption of Fig.~\ref{fig:ASModule}, $D'$ is the intermediate channel numbers.}
		\label{fig:fig3NLC}
		\vspace{-0.3cm}
	\end{figure}
	\subsubsection{Point Nonlocal Cell} 
	Inspired by nonlocal neural networks~\cite{Nonlocal} in 2D images for long-range dependencies learning, we design a specific point nonlocal (PNL) cell for global context aggregation (Fig.~\ref{fig:fig3NLC}). There are two main differences between our point nonlocal cell and component proposed in~\cite{Nonlocal}: (1) We use our sampled points as query points to calculate similarity with entire points in certain layers (say, key points $\mathcal{P}_{\text{k}}$). Furthermore, our query points are not limited within a subset of input point clouds, as each sampled point adaptively updates its coordinate and features by the AS module (Sec.~\ref{AS}). (2) Our output channel is gradually increased with the down-sampling operation in each layer, which avoids information loss in the down-sampling encoder. Specifically, similar with Eq.~\ref{eqattention}, given query point $x_i$ and key point from $\mathcal{P}_{\text{k}}$, the nonlocal operation $\mathcal{NL}$ is defined as:
	\begin{equation}
	\mathcal{NL}(x_i, \mathcal{P}_{\text{k}}) :=  \mathcal{A}(\mathcal{R}(f_i ,f_j)\gamma(f_j), \forall x_j \in \mathcal{P}_{\text{k}}),
	\label{Nonlocal}
	\end{equation}
	where $\mathcal{P}_{k}\in \mathbb{R}^{N\times3}$ stands for the entire $N$ key points in a certain layer. Finally, a single nonlinear convolution layer  $\sigma$ fuse the global context and adjust the channel of each point to the same dimension with the output of PL $D_{l+1}$ (Eq.~\ref{eq1}). Hence, for a sampled point $x_i$,  its updated feature is computed by PNL with function
	\begin{equation}
	f_i^{nl} =  \sigma(\mathcal{NL}(x_i, \mathcal{P}_{k})).
	\end{equation}

	\subsubsection{Local-Nonlocal (L-NL) Fusion} 
	
	By combining PL and PNL, we construct a local-nonlocal module to encode local and global features simultaneously. As shown in Fig.~\ref{fig:ASModule} (b), it uses query points and key points as inputs, and exploit k-NN grouping for neighborhoods searching for each query point. Then, the group coordinates and features of each local region are sent through PL for local context encoding. For PNL, it uses whole key points  to integrate global information for each query point via an attention mechanism.  Finally, for each updated point, a channel-wise sum with a nonlinear convolution $\sigma$ is used to fuse local and global information.

	\subsection{PointASNL}
	\label{PNL}
	By combining the two components proposed in Sec.\ref{AS} and Sec \ref{NL} in each layer, we can implement a hierarchical architecture for both classification and segmentation tasks. 
	
	For the {classification}, we designed a three-layer network and down-sample input points at two levels. In particular,  the first two layers sample 512 and 124 points. The third layer concatenate global features of former two layers with max pooling, where new features are processed by fully connected layers, dropout, and softmax layer, respectively. The batch normalization layers and the ReLU function are used in each layer. Furthermore, skip connections~\cite{resnet} are used in the first two layers.
	
\begin{figure}[t]
		\begin{center}
			\includegraphics[width=\columnwidth]{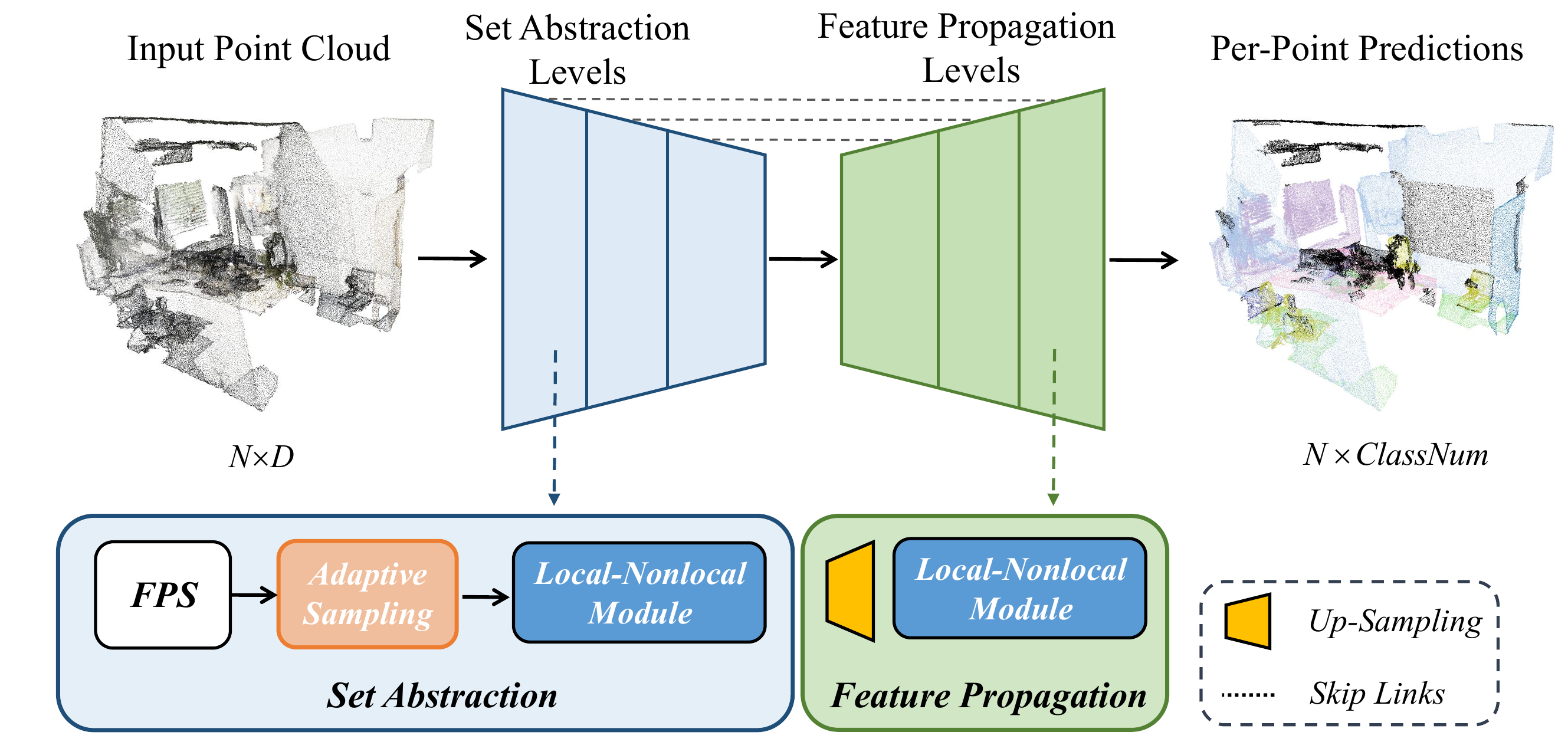}
		\end{center}
		\caption{{Architecture of our PointASNL for point cloud semantic segmentation.} The L-NL modules are used in both encoder and decoder.}
		\label{fig:figure4}
		\vspace{-0.3cm}
	\end{figure}

	For the {segmentation} (see Fig.~\ref{fig:figure4}), each encoder layer is similar with the setting in classification, but network has a deeper structure (1024-256-64-16). In the decoder part, we use 3-nearest interpolation~\cite{pointnet2} to get the up-sampled features and also use the L-NL Block for better feature learning. Furthermore, skip connections are used to pass the features between intermediate layers of the encoder and the decoder.

	\section{Experiment}
	\label{exp}
	
	We evaluate our PointASNL on various tasks, including synthetic dataset, large-scale indoor and outdoor scene segmentation dataset. In all experiments, we implement the models with Tensorflow on one GTX 1080Ti GPU. 
	
	\subsection{Classification}
	We evaluate our model on synthetic dataset \textit{ModelNet10} and \textit{ModelNet40}~\cite{modelnet} for classification, where \textit {ModelNet40} is composed of 9843 train models and 2468 test models in 40 classes and  \textit {ModelNet10} is a subset of \textit{ModelNet40} that consists of 10 classes with 3991 training and 908 testing objects. 

	{\noindent\bf Shape Classification.}  Training and testing data in classification are provided by \cite{pointnet}. For training, we select 1024 points as the input. The augmentation strategy includes the following components: random anisotropic scaling in range $[-0.8, 1.25]$, translation in the range $[-0.1, 0.1]$, and random dropout $20\%$ points. For testing, similar to \cite{pointnet,pointnet2}, we apply voting test using random scaling and then average the predictions. 	In Tab.~\ref{tab:tab1}, our method outperforms almost all state-of-the-art methods in 1024 input points except RS-CNN. Note that RS-CNN~\cite{rscnn} can achieve 93.6\% from 92.9\% on uniformly sampling with tricky voting strategy (the best of 300 repeated tests), which is different from normal random sampling and once voting setting.

	\begin{table}
		\small
		\caption{Overall accuracy on ModelNet10 (M10) and ModelNet40 (M40) datasets. ``pnt'' stands for coordinates of point and ``nor'' stands for normal vector.}
		\begin{center}
			\begin{tabular}{lcccc}
				\hline
				Method &input&\#points &\textit {M10}  &\textit {M40} \\
				\hline
				\hline
				
				O-CNN~\cite{wang2017cnn} &pnt, nor&-&-  &90.6 \\
				SO-Net~\cite{So-net} &pnt, nor&2k&94.1  &90.9 \\
				Kd-Net~\cite{KD-Tree} &pnt&32k&  94.0& 91.8 \\
				PointNet++~\cite{pointnet2} &pnt, nor&5k& - & 91.9  \\
				SpiderCNN~\cite{Spidercnn} &pnt, nor &5k& -&  92.4 \\
				KPConv~\cite{KPCONV}  &pnt&7k& -&  92.9 \\
				
				SO-Net~\cite{So-net} &pnt, nor&5k&\textbf{95.7}  &\textbf{93.4} \\
				
				\hline
				Pointwise CNN~\cite{Pointwise}  &pnt&1k& -  &86.1\\
				ECC~\cite{ecc} &graphs&1k&  90.8&  87.4 \\
				PointNet~\cite{pointnet} &pnt&1k&-  &89.2 \\
				PAT~\cite{Gumbel} &pnt, nor&1k&-  &91.7 \\
				Spec-GCN~\cite{SpecGCN} &pnt&1k&-  &91.8 \\
				PointGrid~\cite{pointgrid} &pnt&1k& - &92.0 \\
				PointCNN~\cite{PointCNN} &pnt&1k&  - &92.2 \\
				DGCNN~\cite{DGCNN} &pnt&1k&  -&  92.2 \\
				PCNN~\cite{PCNN} &pnt&1k& 94.9&  92.3 \\
				PointConv~\cite{PointConv} &pnt, nor&1k& - &92.5 \\
				A-CNN~\cite{acnn} &pnt, nor&1k& 95.5 &92.6 \\
				Point2Sequence~\cite{Point2Sequence} &pnt&1k& 95.3& 92.6 \\
				RS-CNN~\cite{rscnn} &pnt&1k& - & 93.6 \\
				\hline
				PointASNL &pnt&1k&  {95.7}&{92.9} \\
				PointASNL &pnt, nor&1k&  \bf{95.9}&\bf{93.2} \\
				\hline
			\end{tabular}
		\end{center}
		
		\label{tab:tab1}
		\vspace{-0.6cm}
	\end{table}

\begin{figure}[!bp]
		\begin{center}
			\includegraphics[width=1 \linewidth]{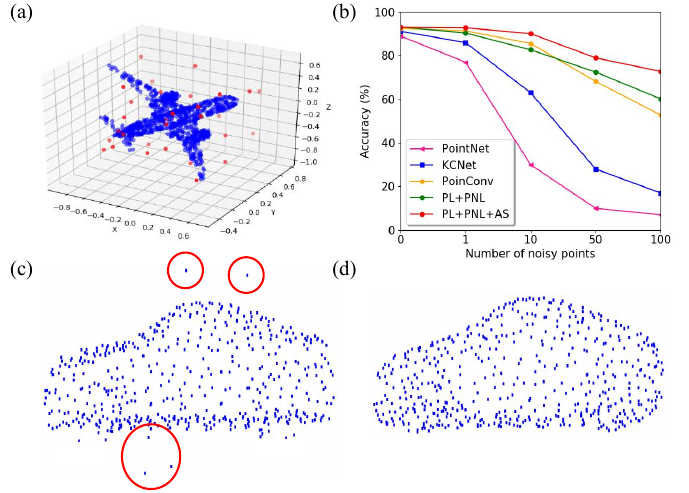}
		\end{center}
		\caption{(a) Point cloud with some points being replaced with random noise. (b) Classification results of different models with noisy points, where PL, PNL, AS mean point local cell, point nonlocal cell and adaptive sampling, respectively. (c) Farthest point sampling on noisy data. (d) Adaptive sampling on noisy data, which maintain the distribution of the point cloud.}
		\label{fig:figure8}
		
	\end{figure}

	{\noindent\bf Shape Classification with Noise.} 
	Most of the methods can achieve decent performance on synthetic datasets, as they have stable distribution and do not contain any noise. Though, such a good performance often leads to a lack of robustness of the model. To further verify the robustness of our model, we did the experiments like KC-Net~\cite{KC-Net} to replace a certain number of randomly picked points with random noise ranging $[-1.0, 1.0]$ during testing. The comparisons with PointNet~\cite{pointnet}, PointConv~\cite{PointConv} and KC-Net~\cite{KC-Net} are shown in Fig.~\ref{fig:figure8} (b). As shown in this figure, our model is very robust to noise, especially after adding the AS module. It can be seen from (c) and (d) that the adaptive sampling guarantees the proper shape of the sampled point clouds, making the model more robust.
	
	
	\subsection{Segmentation}
	{\noindent\bf Indoor Scene Segmentation.}\footnote {Supplementary material shows that with more sampled points and deeper structure, our PointASNL can still achieve further improvement to 66.6\% on \textit{ScanNet} benchmark.} Unlike classification on synthetic datasets~\cite{modelnet,shapenet}, indoor 3D scene segmentation is a more difficult task, because it is real-world point clouds and contains lots of outliers and noise. We use \textit{Stanford 3D Large-Scale Indoor Spaces (S3DIS)}~\cite{s3dis} and \textit{ScanNet v2} (\textit{ScanNet})~\cite{scannet} datasets to evaluate our model.
	
	\textit{S3DIS} dataset is sampled from 3 different buildings, which includes 6 large-scale indoor areas with 271 rooms.  Each point in this dataset has a semantic label that belongs to one of the 13 categories. We compare mean per-class IoU (mIoU) on both 6-fold cross-validation over all six areas and Area 5 (see supplementary material). \textit{ScanNet} dataset contains 1513 scanned indoor point clouds for training and 100 test scans with all semantic labels unavailable. Each point has been labeled with one of the 21 categories. We submitted our results to the official evaluation server to compare against other state-of-the-art methods on the benchmark.

	\begin{table}[!tbp]
		\centering
		\small
		\caption{Segmentation results on indoor S3DIS and ScanNet datasets in mean per-class IoU (mIoU,\%).}\label{tab:tabindoor}
		\vspace{0.2cm}
		\setlength{\tabcolsep}{5mm}{
			\begin{tabular}{l|c|c}
				\hline
				{Method}    & {S3DIS}         & {ScanNet}  \\       \hline \hline       
				\multicolumn{3}{c}{\textit {methods use unspecific number of points as input}}\\\hline
				{TangentConv~\cite{TangentConv}} & {52.8}          & 40.9                 \\        
				{SPGraph~\cite{SPGraph}}     & {62.1}          & -                              \\ 
				{KPConv~\cite{KPCONV}}      & {\textbf{70.6}} & \textbf{68.4}                           \\ \hline
				
				\multicolumn{3}{c}{\textit {methods use fixed number of points as input}}\\\hline
				{PointNet++~\cite{pointnet2}}  & {53.4}             & 33.9      \\
				{DGCNN~\cite{DGCNN}}       & {56.1}          & -       \\ 
				{RSNet~\cite{RSNet}}       & {56.5}          & -         \\ 
				{PAT~\cite{Gumbel}}         & {64.3}          & -       \\ 
				{PointCNN~\cite{PointCNN}}    & {65.4}          & 45.8     \\ 
				{PointWeb~\cite{PointWeb}}    & {66.7}          & -      \\ 
				
				{PointConv~\cite{PointConv}}   & {-}             & 55.6  \\ 
				{HPEIN~\cite{HPEIN}}       & {67.8}          & 61.8    \\  \hline
				{PointASNL}        & {\textbf{68.7}} & \textbf{63.0}  \\ \hline
				\multicolumn{3}{c}{ }
		\end{tabular}}
		\vspace{-0.6cm}
	\end{table}
	
	\begin{figure*}
		\begin{center}
			\includegraphics[width=0.95 \linewidth, height=5.6cm]{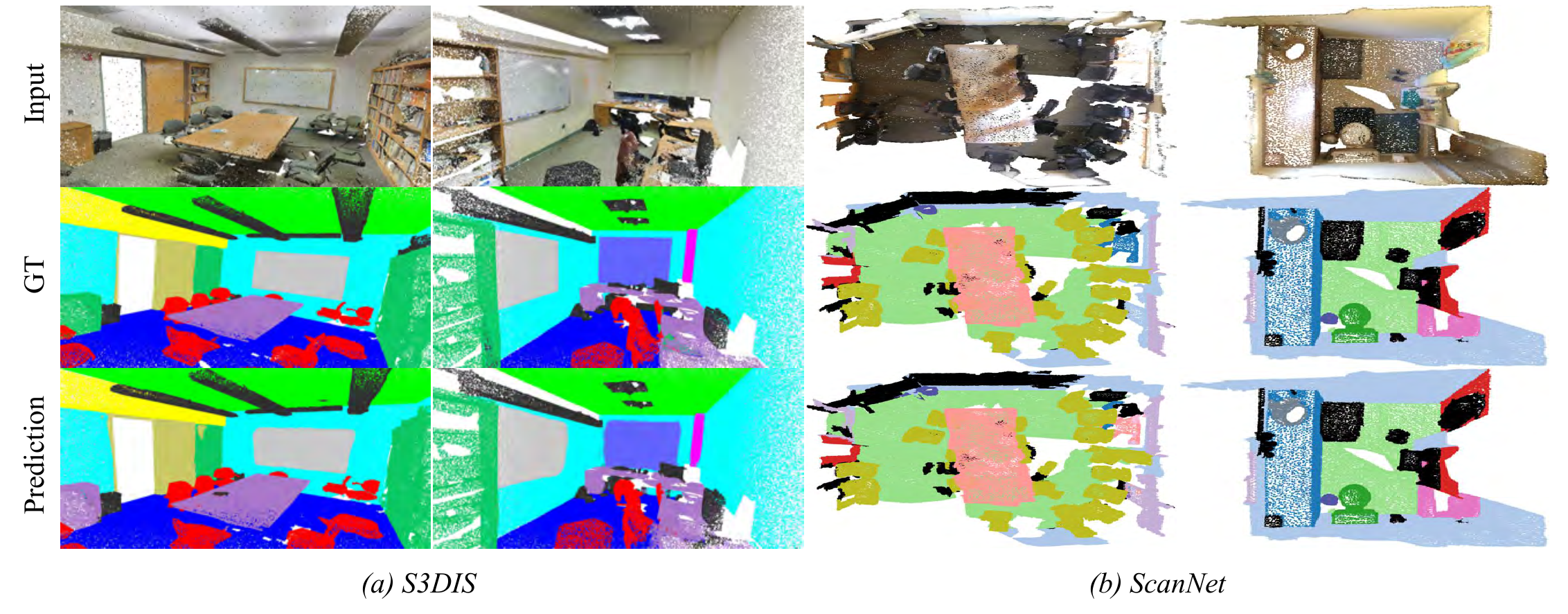}%
		\end{center}
		\vspace{-0.5cm}
		\caption{Examples of indoor semantic segmentation on \textit{S3DIS} and \textit{ScanNet} datasets.}
		\label{fig:figindoor}
		
	\end{figure*}
	
	During the training process, we generate training data by randomly sample $1.5m\times1.5m\times3m$ cubes with 8192 points from the indoor rooms. $0.1m$ padding of sampled cubes is used to increase the stability of the cube edge prediction, which is not considered in the loss calculation. On both datasets, we use points position and RGB information as features. We did not use the relative position in PointNet~\cite{pointnet} as a feature to train the model in \textit{S3DIS}, because our model already learns relative position information well. During the evaluation process, we use a sliding window over the entire rooms with $0.5m$ stride to complement $5$ voting test. 
	
	In Tab.~\ref{tab:tabindoor}, we compare our PointASNL with other state-of-the-art methods under the same training and testing strategy (randomly chopping cubes with a fixed number of points), e.g., PointNet++~\cite{pointnet2}, PointCNN~\cite{PointCNN},  PointConv~\cite{PointConv}, PointWeb~\cite{PointWeb} and HPEIN~\cite{HPEIN}. We also list results of another kind of methods (using points of unfixed number or entire scene as input),  e.g., {TangentConv~\cite{TangentConv}} and {KPconv~\cite{KPCONV}}. All methods use only point clouds as input without voxelization.

	As shown in Tab.~\ref{tab:tabindoor}, PointASNL outperforms all methods with same the training strategy in both \textit{S3DIS} and \textit{ScanNet}. In particular, our result is 8\% higher than previous state-of-the-art PointConv~\cite{PointConv} on the ScanNet with the same experiment setting, in which the convolution design is similar to our PL cell. Nevertheless, without proper sampling and global information support, it cannot achieve such results with the same network architecture.
	
	On the other hand, training using more points as input can obtain more information. Instead of learning from randomly selected cubes with fixed number,  KP-Conv~\cite{KPCONV} performs grid sampling based on the assumption of local label consistency so that larger shape of the point cloud can be included as input. 
	
	The qualitative results are visualized in Fig.~\ref{fig:figindoor}. Our method can correctly segments objects even in complex scenes.
	
	{\noindent\bf Outdoor Scene Segmentation.} Compared with its indoor counterpart, an outdoor point cloud covers a wider area and has a relatively sparser point distribution with noise. For this reason, it is more challenging to inference from outdoor scenes. 
	
	\begin{table}[]
		\centering
		\small
		\setlength{\tabcolsep}{4mm}{
			\caption{{Semantic Segmentation results on SemanticKITTI, where “pnt” stands for coordinates of point and SPGraph~\cite{SPGraph} uses all of points as input.}}
			\vspace{0.2cm}
			\label{tab:tabkitti}
			\begin{tabular}{l|c|c}
				\hline
				{Method}&  input  & mIoU(\%)  \\ 
				\hline \hline
				
				{PointNet~\cite{pointnet}} &50k pnt     & 14.6    \\ 
				{SPGraph~\cite{SPGraph}}  &-     & 17.4  \\ 
				{SPLATNet~\cite{su2018splatnet}} &50k pnt      & 18.4     \\ 
				{PointNet++~\cite{pointnet2}{}} &45k pnt    & 20.1     \\ 
				{TangentConv~\cite{TangentConv}} &120k pnt   & {40.9}   \\
				\hline
				{PointASNL} &8k pnt& \textbf{46.8}          \\ \hline
		\end{tabular}}
		\vspace{-0.5cm}
	\end{table}

	\begin{figure}[!hbp]
		\vspace{-0.5cm}
		\begin{center}
			\includegraphics[width=1 \linewidth]{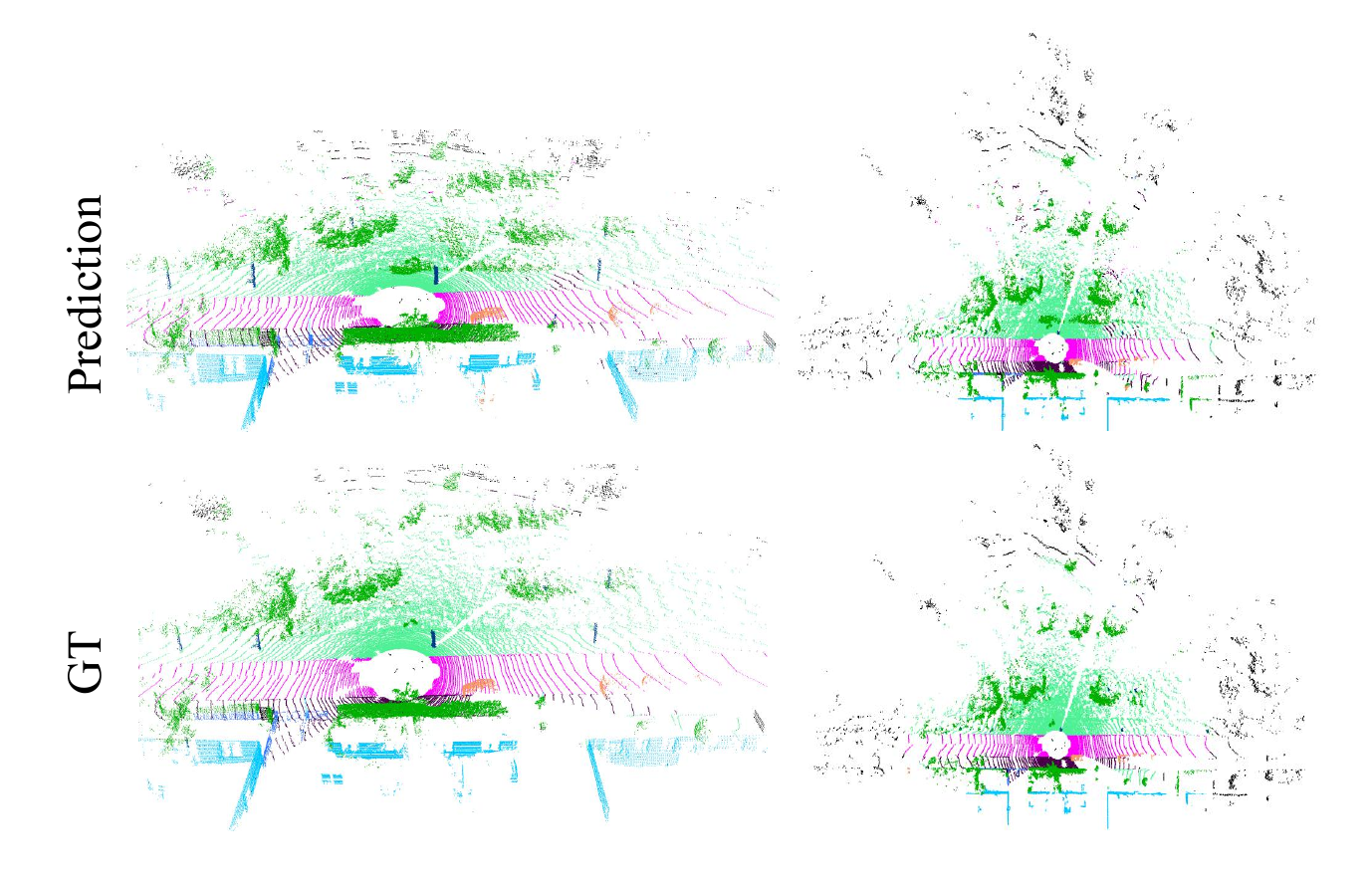}
		\end{center}
		\caption{ {Example of outdoor SemanticKITTI datasets}. }
		\label{fig:SemanticKITTI}
	\end{figure}
	
	We evaluated our model on SemanticKITTI~\cite{behley2019iccv}, which is a large-scale outdoor scene dataset, including 43,552 scans captured in the wild. The dataset consists of 22 sequences (00 to 10 as the training set, and 11 to 21 as the test set), each of which contains a series of sequential laser-scans. Each individual scan is a point clouds generated with a commonly used automotive LiDAR. The whole sequence can be generated by aggregating multiple consecutive scans.
	
	In our experiments, we only evaluated our model under a single scan semantic segmentation. In the single scan experiment~\cite{behley2019iccv}, sequential relations among scans in the same sequence are not considered. The total number of 19 classes is used for training and evaluation. Specifically, the input data generated from the scan is a list of coordinates of the three-dimensional points along with their remissions. 
	
	During training and testing, we use a similar sliding windows based strategy as indoor segmentation. Since point clouds in the outdoor scene are more sparse, we set the size of the cube with $10m\times10m\times6m$  and $1m$ padding. In Tab.~\ref{tab:tabkitti}, we compare PointASNL with other state-of-the-art methods. Our approach outperforms others by a large margin. Supplementary material shows that our method achieves the best result in 13 of 19 categories. Furthermore, Fig.~\ref{fig:SemanticKITTI} illustrates our qualitative visualization of two samples, even if the scene is covered with a lot of noise caused by unmanned collection, our model can still predict perfectly.

	



	\subsection{Ablation Study}
	To further illustrate the effectiveness of proposed AS and L-NL module, we designed an ablation study on both the shape classification and the semantic segmentation. The results of the ablation study are summarized in Tab.~\ref{tab:tab3}.  
	
	We set two baselines: A and B. Model A only encodes global features by PNL, and model B only encodes local features. The baseline model A gets a low accuracy of 90.1\% and 45.7\% IoU on segmentation, and model B gets 92.0\% and 56.1\%, respectively. When we combine local and global information (models C), there is a notable improvement in both classification and segmentation. Finally, when we add the AS module, the model will have a significant improvement in the segmentation task (93.2\% and 63.5\% in model D). 
	
	\begin{table}
		\small
		\renewcommand\tabcolsep{3.5pt} 
		\caption{Ablation study on \textit {ModelNet40} and \textit{ScanNet v2} validation set. PL, PNL and AS mean point local cell, point nonlocal cell and adaptive sampling.}
		\begin{center}
			\begin{tabular}{cc|cc}
				
				Model& Ablation  & \textit {ModelNet40}& \textit{ScanNet} \\
				\hline
				\hline 
				A& PNL only & {90.1}& {45.7} \\	
				B& PL only  & {92.0}& {56.1} \\
				
				C& PL+PNL & {93.2}& {60.8} \\	
				D&PL+PNL+AS & {\bf{93.2}}& {\bf{63.5}} \\	
				\hline
				E& PointNet2~\cite{pointnet} & {90.9}& {48.9} \\			
				F& PointNet2+PNL & {93.0}& {54.6} \\			
				G& PointNet2+PNL+AS & {92.8}& {55.4} \\\hline
				
				H&DGCNN~\cite{DGCNN}& {92.2}& {52.7} \\
				I&DGCNN+PNL& {92.9}& {56.7} \\
				J&DGCNN+PNL+AS & {93.1}& {58.3} \\
				
				\multicolumn{3}{c}{ }\\
				
			\end{tabular}
			\label{tab:tab3}
		\end{center}
		\vspace{-1cm}
	\end{table}
	Furthermore, our proposed components L-NL module and AS module can directly improve the performance of other architecture. When we use PointNet++~\cite{pointnet2} in our PL cell (model F), it will reduce the error of classification and segmentation tasks by 23.1\% and 12.6\%, respectively, with its original model (model E). It should be noted that the AS module does not increase the accuracy of the classification task, even reduced the accuracy of classification when adding on PointNet++ (model F). This is because the synthetic dataset does not have a lot of noise like scene segmentation, for some simpler local aggregation (e.g., max pool), it may make them unable to adapt the uneven point cloud distribution after using AS.  Furthermore, we also use DGCNN~\cite{DGCNN} as our local aggregation baseline (model H), and fused architecture (model I and J) can largely improve the performance on two datasets. 
	
	\subsection{Robustness for Sparser Point Clouds}
	\begin{figure}[!tbp]
		\begin{center}
			\includegraphics[width=\linewidth]{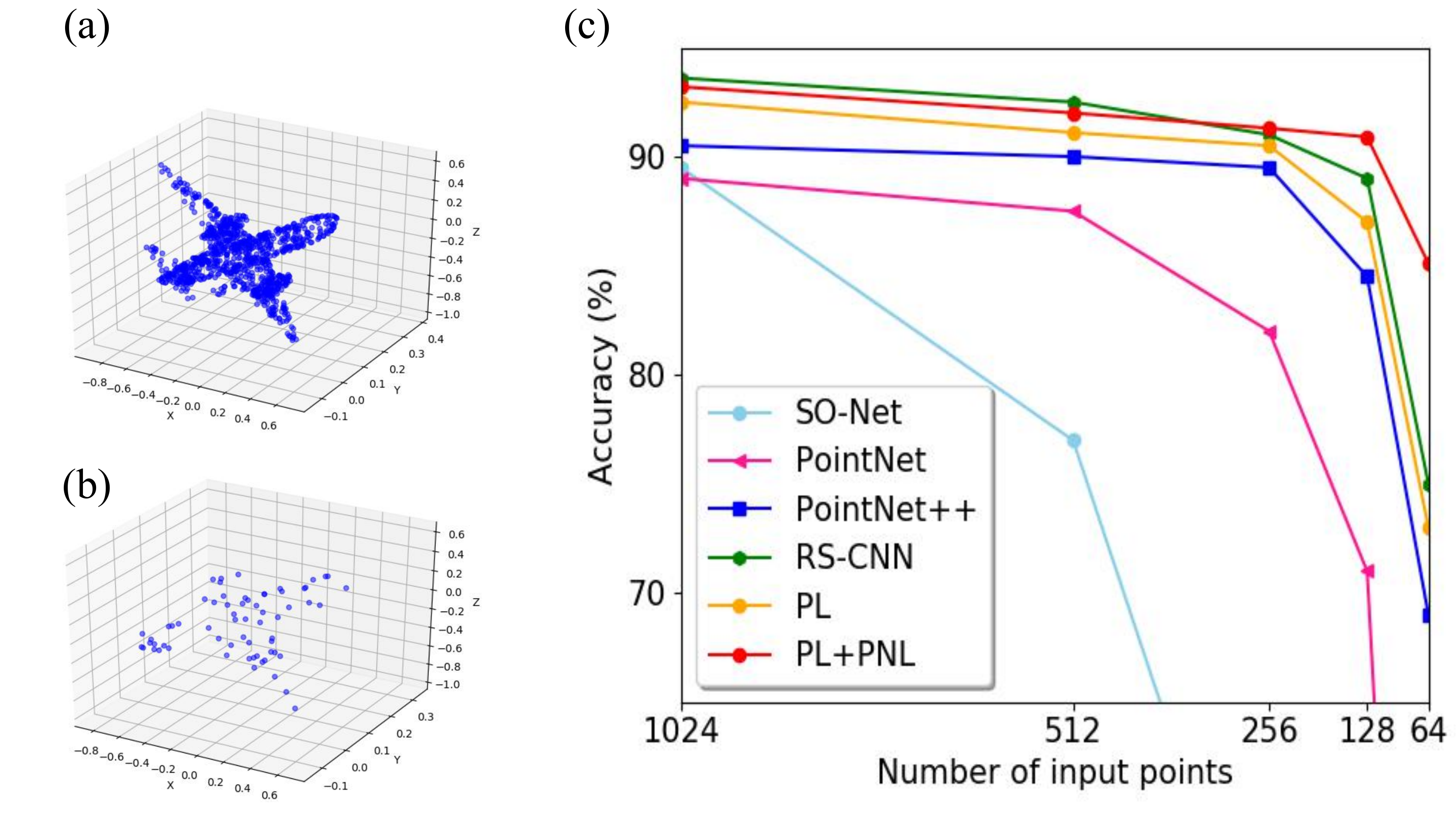}
		\end{center}
		\caption{ { (a) A sample of input point cloud. (b)  A sample of input point cloud after randomly select 64 points (c) Results of testing with sparser points}. }
		\label{fig:sp_fig1}
		\vspace{-0.4cm}
	\end{figure}
	To further verify the robustness of the PointASNL model, we take sparser points (i.e., 1024, 512, 256, 128 and 64) as the input to various models trained with 1024 points. Then we compare our method with PointNet~\cite{pointnet}, PointNet++~\cite{pointnet2}, SO-Net~\cite{So-net} and the recent state-of-the-art RS-CNN~\cite{rscnn}. We follow these methods to apply random input dropout during the training.

	As can be seen from the Fig~.\ref{fig:sp_fig1} (c), PNL can greatly improve the robustness of our model with different density inputs. In particular, when the input contains only 64 points, PNL can even help to improve the accuracy of our model, from 73.9\% to 85.2\%, which largely exceeds the current state-of-the-art RS-CNN~\cite{rscnn} (about 75\%). The experimental results fully demonstrate that the use of local and global learning methods can greatly improve the robustness of the model.  As shown in Fig~.\ref{fig:sp_fig1} (a) and (b), when the input points reduce to 64, even humans can hardly recognize the airplane, but our model can classify it correctly. Such superior robustness makes our proposed PointASNL model suitable for raw noisy point clouds with limited sampling points, especially for large scale outdoor scenario.

	\section{Conclusion}
	We have presented the adaptive sampling (AS) and the local-nonlocal (L-NL) module to construct the architecture of PointASNL for robust 3D point cloud processing. By combining local neighbors and global context interaction, we improve traditional methods dramatically on several benchmarks. Furthermore, adaptive sampling is a differentiable sampling strategy to fine-tune the spatial distribution of sampled points, largely improve the robustness of the network. Experiments with our state-of-the-art results on competitive datasets and further analysis illustrate the effectiveness and rationality of our PointASNL.\\

	{\noindent{\bf Acknowledgment.} The work was supported by grants No. 2018YFB1800800, NSFC-61902335, No. 2019E0012, No. ZDSYS201707251409055, No. 2017ZT07X152, No. 2018B030338001, and CCF-Tencent Open Fund.}
	
	{\footnotesize
		\bibliographystyle{ieee_fullname}
		\bibliography{egbib}
	}
	
	\newpage
	
	\begin{center}
		{\textit{\Large\bf Supplementary Material}}
	\end{center}
	
	\maketitle
	
	\thispagestyle{empty}
	\setcounter{section}{0}
	\setcounter{figure}{0}
	\setcounter{table}{0}
	\renewcommand\thesection{\Alph{section}}

	\section{Overview}
	In this supplementary material, we first provide more additional experiments to further verify the superiority of our model in Section~\ref{ae}. Besides, we show the our network architecture details in Section~\ref{na}.

	\section{Additional Experiment}
	\label{ae}
	\begin{figure}[b]
		\begin{center}
			\includegraphics[width=\linewidth,height=5.5cm]{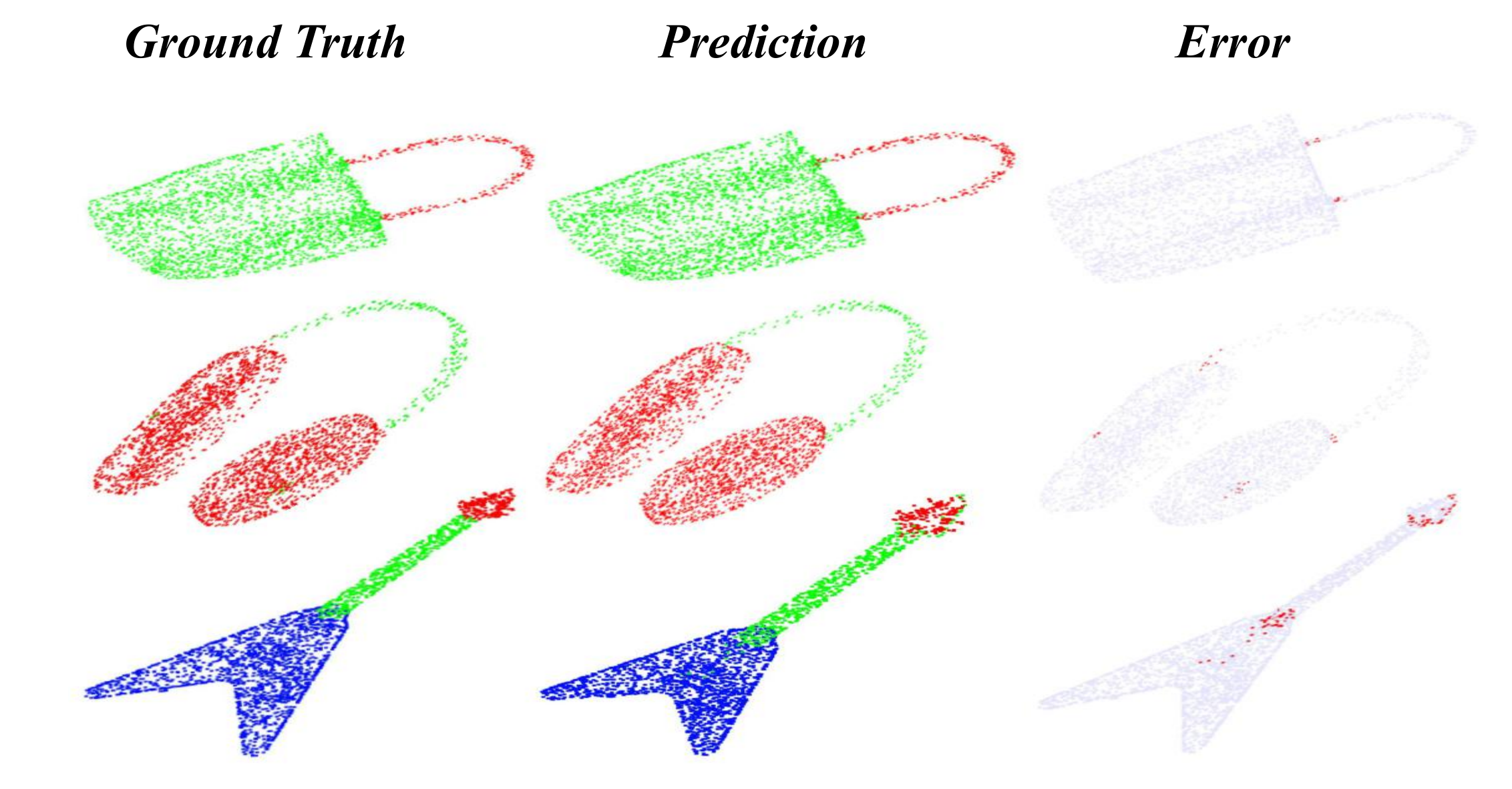}
		\end{center}
		\caption{ {Selected results of part segmentation}.}
		\label{fig:sp_fig2}
	\end{figure}
	\subsection{Part Segmentation}
	Due to space limitation, we illustrate the part segmentation experiments using man-made synthetic dataset ShapeNet~\cite{shapenet}, which contains 16,881 shapes from 16 classes and 50 parts. We use the data provided by \cite{pointnet2} and adopt the same training and test strategy, i.e., randomly pick 2048 points as the input and concatenate the one-hot encoding of the object label to the last layer.

	The quantitative comparisons with the state-of-the-art point-based methods are summarized in Tab.~\ref{tab:ShapeNet}. Note that we only compare with methods use 2048 points. When compared with the state-of-the-arts, PointASNL achieves comparable result, which is only slightly lower than RS-CNN~\cite{rscnn} using different sampling and voting strategy (as the same reason for classification task).

	\subsection{Selection of Adaptive Sampling}
	Two variable conditions, i.e., the sampling strategy for initial sampled points and deformation method, are investigated for this issue. Tab.~\ref{as_compare} summarizes the results. For the initial sampling points, we chose two strategies, i.e., FPS and random sampling (RS). Also for local coordinate points and feature updates, we compare the effects of using the weight learning by group feature (GF) and simple average of all neighbors’ coordinates and features. Note that the number of neighbors is set to be equal for a fair comparison.
	
	As Tab.~\ref{as_compare} shows,, if we just use RS sample the initial points  and then average their coordinates and features (model A), we will get very low accuracy of 87.9\%. However, if we use FPS instead of RS (model B), it can increase to 91.5\%. Furthermore, model C and D illustrate the weight learning using group features can largely increase the inference ability of our model.  However, if we use RS as sampling strategy, it will cause some accuracy loss while we add the group features learning.  This shows that AS module can only finely adjust the distribution of the sampled point cloud instead of 'creating' the missing information.
	\begin{table}[t]
		\small
		\begin{center}
			\caption{The results (\%) of four selection strategies on adaptive sampling.  For a fair comparison, the number of neighbors is set to be equal in each layer between the two models.}
				\vspace{0.1cm}
			\begin{tabular}{c|cc|cc|c}
				\hline
				Model& RS& FPS& Average & GF & \textit {ModelNet40}\\
				\hline
				\hline 
				
				A& \checkmark&&\checkmark& & {87.9} \\	
				B& &\checkmark&\checkmark& & {91.5} \\
				
				C& \checkmark&&& \checkmark& {92.3} \\	
				D&&\checkmark&&\checkmark & {\bf{93.2}} \\	
				\hline
			\end{tabular}
			\label{as_compare}
		\end{center}
	\end{table}

	\subsection{Visualization of L-NL Module}
	
\begin{figure}[!tbp]
		\begin{center}
			\includegraphics[width= 0.90\linewidth]{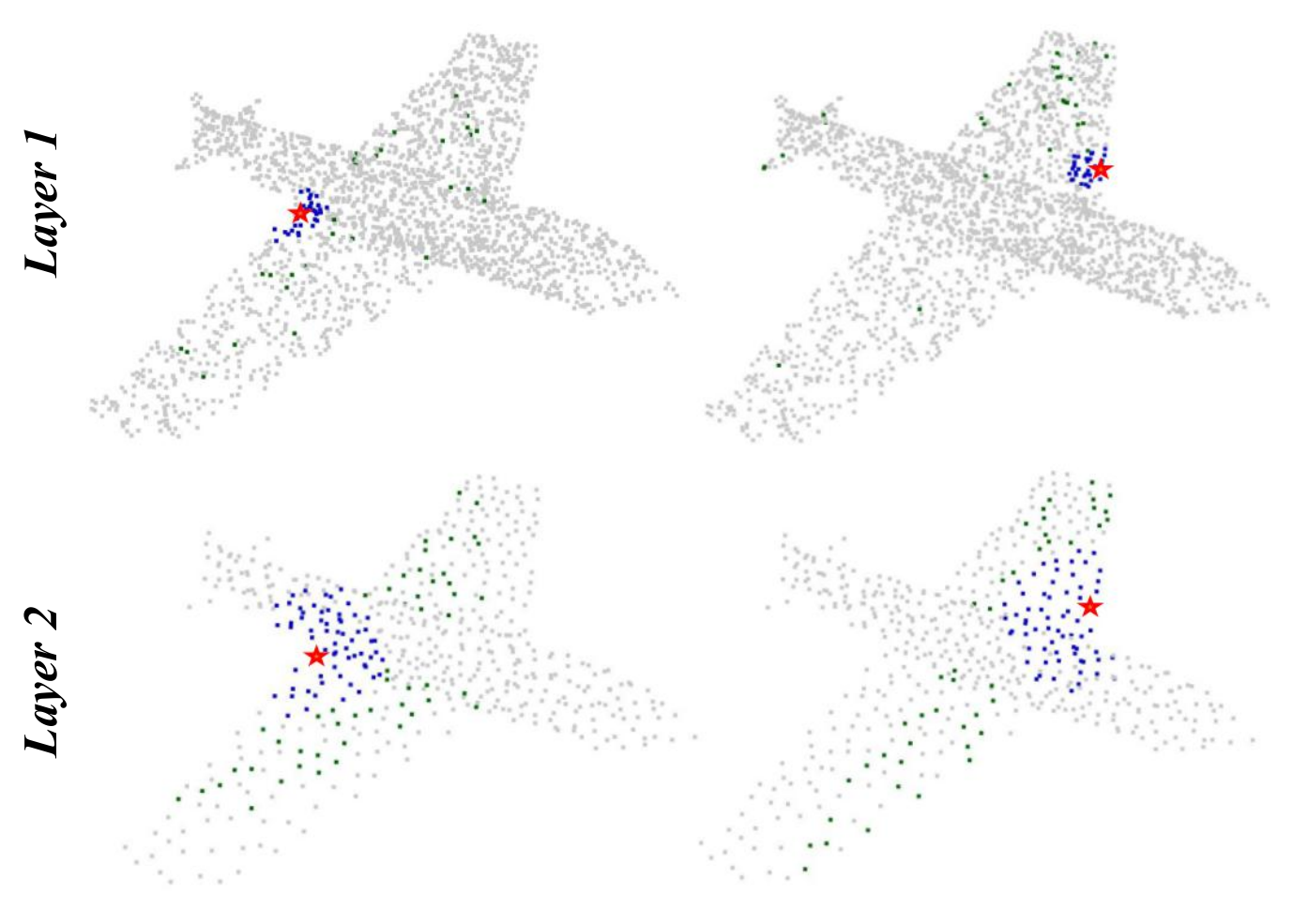}
		\end{center}
		\caption{ {Visualization of local-global learning}. For each sampled point (red), we search its local neighbors (blue) and the K points with the highest global response value (green), where K is equal to the number of local neighborhoods. }
		\vspace{-0.5cm}
		\label{fig:figure9}
	\end{figure}

	We further demonstrate the local-global learning of PointASNL in Fig.~\ref{fig:figure9}. In the first layer of the network, PNL can find global points that have similar characteristics with sampled points (e.g., edge and normal vectors). In the second layer, these global highly responsive points have the same semantics information with sampled points, even when sampled points are at the junction of the two different semantics. This is why global features can help sampled points to better aggregate local features.
	
	\subsection{Visualization of Adaptive Sampling}
	When the input point cloud has a lot of noise, adaptive sampling has the ability to ensure the distribution of the sample point manifold. We give some examples of comparative visualization in Fig.~\ref{fig:as} to prove the robustness of the AS module. As can be seen from Fig.~\ref{fig:as}, AS module can effectively reduce noise in the sample points and maintain the shape of the sampled manifold.
	
	\subsection{Further Improvement of PointASNL}
	The result in manuscript only conducts a {fair comparison} (same model structure and training strategy) against appealing recent methods under the same setting of PointNet++~\cite{pointnet2}. However, our PointASNL can still achieve further improvement if we use other data pre-processing or deeper structure. 
	
	\begin{table}[]
			\small
			\renewcommand\tabcolsep{5.5pt} 
			\begin{center}
				\caption{The results (mIoU, (\%)) on \textit{ScanNet} v2 validation set with other model setting.}
					\vspace{0.1cm}
				\begin{tabular}{c|cccc}
					\hline
					Model& input & grid sample& deeper & mIoU \\
					\hline
					\hline 
	
					A& 8192 pnt&&&  {63.5} \\	
					B& 8192 pnt&\checkmark&&  {64.5} \\
	
					C& 10240 pnt&\checkmark&&  {64.8} \\	
					D&10240 pnt&\checkmark&\checkmark & {\bf{66.4}} \\	
					\hline
				\end{tabular}
				\label{scannet_abl}
			\end{center}
			\vspace{-0.5cm}
	\end{table}
	As shown in Tab.~\ref{scannet_abl}, our PointASNL can still improve its performance if we use grid sampling pre-processing, more input points and deeper structure. As for the structure of deeper PointASNL, we add an additional point local cell at the end of each layer. Furthermore, by conducting ensemble learning with model from different training epochs, we can finally achieve 66.6\% on \textit{ScanNet} benchmark.
	
	\subsection{Concrete Results}
	In this section, we give our detailed results on the S3DIS (Tab.\ref{tab:S3DIS_A5} and Tab.\ref{tab:S3DIS_6F}) and SemanticKITTI (Tab.\ref{tab:KITTI}) dataset as a benchmark for future work. \textit{ScanNet}~\cite{scannet} is an online benchmark, the class scores can be found on its website. Furthermore, we provide more visualization results to illustrate the performance of our model in complicated scenes.

	\section{Network Architectures}
	\label{na}
	\subsection{Layer Setting}
	For each encoder layer, it can be written as the following form: \textit{Abstraction(npoint, nsample, as\_neighbor, mlp)}, where \textit{npoint} is the number of sampled points of layer. \textit{nsample} and \textit{as\_neighbor} are number of group neighbors in point local cell and AS module, and they share the same k-NN query. \textit{mlp} is a list for MLP construction in our layers and used in both PL and PNL. Tab.~\ref{config} shows the configuration of PointASNL on both classification and segmenttaion tasks.
	
	\begin{table}[t]
		\small
		\renewcommand\tabcolsep{5.5pt} 
		\begin{center}
			\caption{Network Configurations.}
			\begin{tabular}{c|cccc}
				\hline
				Layer& npoint& nsample& as\_neighbor & mlp\\
				\hline
				\hline 
				Task		&	\multicolumn{4}{c}{\textit {Classification}}\\\hline
				1& 512&32&12&[64,64,128] \\	
				2& 128&64&12&[128,128,256]  \\
				3& 1&-&-&[256,512,1024]  \\\hline
				Task		&	\multicolumn{4}{c}{\textit {Segmentation}}\\\hline
				1& 1024&32&8&[32,32,64] \\	
				2& 256&32&4&[64,64,128]  \\
				
				3& 64&32&0& [128,128,256] \\	
				4& 36&32&0& [256,256,512] \\	
				\hline
				
			\end{tabular}
			\label{config}
		\end{center}
	\vspace{-0.3cm}
	\end{table}	
	\subsection{Loss Function}
	Like other previous works, we use cross entropy (CE) loss in classification and part segmentation, and consider the number of each category as weights in semantic segmentation. Furthermore, in order to avoid the sampled points being too close to each other in some local areas after the AS module transformation, we also use {{Repulsion Loss}}~\cite{PUNet} to restrict the deformation of sampled point clouds. In particular, we only use this loss in the first layer since it has the highest point density. The Repulsion loss does not bring any performance improvement, but the training procedure is significantly accelerated.
	
	Altogether, we train the PointASNL in an end-to-end manner by minimizing the following joint loss function:
	\begin{equation}
	\begin{split}
	&L(\theta) = L_{{\text{CE}}} + \alpha L_{{\text{Rep}}} + \beta||\theta||^2,\\
	&L_{{\text{Rep}}} = \sum^N_{i=0}\sum_{i'\in N(x_i)} w(||x_{i'} - x_i||), 
	\end{split}
	\end{equation}
	where $\theta$ indicates the parameters in our network, $\alpha= 0.01$ balances the CE loss and Repulsion loss, and $\beta$ denotes the multiplier of the weight decay. For Repulsion loss, it penalizes the sampled point $x_i$ only when it is too close to its neighboring points $x_{i'}\in N(x_i)$. $w(r) = e^{−r^2 /h^2}$ is a fast-decaying weight function and $N$ is the number of sampled points.
	
	The Repulsion loss also ensures that each sample point itself has a larger weight in the AS module in a relatively constant density, which makes them cannot move too far.
	
	\subsection{Model Speeds}
	Tab.~\ref{speed} shows the statistics of our models on different datasets. Since our L-NL module only uses sampled points as query points instead of the whole point cloud, the AS module and NL cell can be both efficient and effective with the bottleneck structures (only around 30\% extra time).

		\begin{table}[b]
			\small
			\renewcommand\tabcolsep{5.5pt} 
			\begin{center}
				\caption{The running time on \textit{ModelNet40} and \textit{ScanNet} datasets.}
				\vspace{0.1cm}
				\begin{tabular}{c|ccc}
					\hline
					Dataset& process&input & time (s/sample) \\
					\hline
					\hline 
	
					\textit{ModelNet40}&Training&1024 pnt& 0.00046 \\	
					\textit{ScanNet}&Training& 8192 pnt&0.17611 \\\hline 
	
					\textit{ModelNet40}&Inference& 1024 pnt& 0.00024 \\	
					\textit{ScanNet}&Inference&8192 pnt& 0.11363 \\	
					\hline
				\end{tabular}
				\label{speed}
			\end{center}
			\vspace{-0.5cm}
	\end{table}

	\begin{table*}[!htp]
		\renewcommand\tabcolsep{2.5pt} 
		\caption{Part segmentation performance with part-avaraged IoU on \textit{ShapeNetPart}.}
		\begin{center}
			\begin{tabular}{l|c|cccccccccccccccc}
				\hline
				
				Method &pIoU &areo& bag& cap& car &chair&ear  & guitar &knife& lamp& laptop &motor &mug &pistol& rocket& skate  &table\\
				& && & &  && phone &  && &  & & && &board   &\\
				\hline
				\#shapes& &2690& 76& 55& 898 &3758&{69} & 787 &392& 1547& 451 &202 &184 &286& 66& 152 &5271\\
				\hline
				PointNet~\cite{pointnet} &83.7& 83.4 &78.7 &82.5 &74.9 &89.6 &73.0 &91.5 &85.9& 80.8 &95.3& 65.2 &93.0 &81.2& 57.9 &72.8&80.6\\
				SO-Net~\cite{So-net}&84.9& 82.8& 77.8& \bf{88.0} &77.3& 90.6& 73.5 &90.7 &83.9 &82.8& 94.8 &69.1 &94.2& 80.9& 53.1& 72.9& 83.0\\
				PointNet++~\cite{pointnet2} &85.1 &82.4 &79.0 &87.7& 77.3 &90.8 &71.8 &91.0& 85.9& 83.7& 95.3 &71.6& 94.1& 81.3& 58.7& 76.4& 82.6\\
				DGCNN~\cite{DGCNN}&85.1 &\bf{84.2} &83.7& 84.4& 77.1 &90.9 &78.5 &91.5& 87.3 &82.9 &96.0 &67.8 &93.3 &82.6& 59.7& 75.5& 82.0\\
				P2Sequence~\cite{Point2Sequence} & 85.2& 82.6 &81.8 &87.5& 77.3 &90.8& 77.1 &91.1 &86.9 &83.9 &95.7 &70.8 &94.6 &79.3& 58.1& 75.2& 82.8\\
				PointCNN~\cite{PointCNN}& 86.1& 84.1& \bf{86.5}& 86.0&\bf{80.8} &90.6& 79.7 &92.3 &\bf{88.4}& 85.3& \bf{96.1} &77.2 &95.2 &\bf{84.2} &\bf{64.2} &\bf{80.0}& 83.0\\
				RS-CNN~\cite{rscnn} &\bf{86.2}& 83.5 &84.8& 88.8 &79.6 &91.2 &\bf{81.1}& \bf{91.6} &88.4& \bf{86.0 }&96.0& 73.7 &94.1 &83.4 &60.5& 77.7& \bf{83.6}\\
				\hline
				PointASNL & {{86.1}}	&84.1&	84.7&	87.9	&79.7&\bf{92.2}	&73.7&	91.0	&87.2&	84.2	&95.8&	\bf{74.4}	&\bf{95.2}&	81.0	&63.0&	76.3	&83.2 \\
				\hline
			\end{tabular}
		\end{center}
		
		\label{tab:ShapeNet}
	\end{table*}
	
	\begin{table*}[!htp]
		\renewcommand\tabcolsep{2.5pt} 
		\caption{Semantic segmentation results on \textit{S3DIS} dataset evaluated on Area 5.}
		\begin{center}
			\begin{tabular}{l|ccc|ccccccccccccc}
				\hline
				
				Method& OA &mAcc& mIoU& ceiling &floor& wall& beam& column& window& door& table&  chair &sofa &bookcase& board& clutter \\
				\hline
				\hline
				
				PointNet~\cite{pointnet} &- &49.0 &41.1&88.8 &97.3& 69.8& \bf{0.1}& 3.9& 46.3& 10.8& 52.6& 58.9 &40.3 &5.9 &26.4& 33.2\\
				
				PointCNN~\cite{PointCNN} &85.9& 63.9 &57.3& 92.3& 98.2& \bf{79.4}& 0.0 &17.6& 22.8& {62.1} &74.4&80.6 &31.7&66.7 &62.1& \bf{56.7}\\
				PointWeb~\cite{PointWeb}&87.0& 66.6& 60.3 &92.0 &\bf{98.5} &79.4& 0.0 &21.1 &59.7 &34.8 &76.3 &\bf{88.3} &46.9&\bf{ 69.3}&64.9 &52.5\\
				HPEIN~\cite{HPEIN} &87.2& {68.3}& {61.9}& 91.5& 98.2 &81.4& 0.0 &23.3 &\bf{65.3} &40.0 &75.5& 87.7 &\bf{58.5} &67.8&\bf{ 65.6} &49.7\\
				\hline
				PointASNL & \bf{87.7}&\bf{68.5} &\bf{62.6}&	\bf{94.3}	&98.4	&79.1	&0.0	&\bf{26.7}	&55.2&	\bf{66.2}&	\bf{83.3}	&86.8&	47.6&	68.3	&56.4	&52.1 \\
				\hline
			\end{tabular}
		\end{center}
		\label{tab:S3DIS_A5}
	\end{table*}
	
	\begin{table*}[!htp]
		\renewcommand\tabcolsep{2.5pt} 
		\caption{Semantic segmentation results on the \textit{S3DIS} dataset with 6-fold cross validation.}
		\begin{center}
			\begin{tabular}{l|ccc|ccccccccccccc}
				\hline
				
				Method& OA &mAcc& mIoU& ceiling &floor& wall& beam& column& window& door& table&  chair &sofa &bookcase& board& clutter \\
				\hline
				\hline
				
				PointNet~\cite{pointnet} &78.5 &66.2 &47.6 &88.0& 88.7 &69.3 &42.4 &23.1& 47.5 &51.6& 42.0 &54.1& 38.2 &9.6& 29.4& 35.2\\
				RSNet~\cite{RSNet}& -&66.5& 56.5 &92.5 &92.8& 78.6& 32.8& 34.4& 51.6 &68.1 &59.7& 60.1 &16.4 &50.2& 44.9& 52.0\\
				A-CNN~\cite{acnn}&87.3&-& 62.9& 92.4 &96.4 &79.2& 59.5 &34.2 &56.3& 65.0& 66.5& \bf{78.0} &28.5 &56.9& 48.0& 56.8\\
				PointCNN~\cite{PointCNN} &88.1 &75.6& 65.4 &94.8& 97.3& 75.8 &\bf{63.3} &\bf{51.7} &58.4 &57.2 &\bf{71.6} &69.1& 39.1& 61.2& 52.2& 58.6\\
				PointWeb~\cite{PointWeb}&87.3 &76.2 &66.7& 93.5& 94.2& 80.8 &52.4& 41.3 &{64.9} &68.1 &71.4 &67.1& \bf{50.3}& \bf{62.7} &{62.2}& 58.5\\
				\hline
				PointASNL & \bf{88.8}&\bf{79.0}&\bf{68.7}&	\bf{95.3}&\bf{97.9}&	\bf{81.9}&	47.0&	48.0&	\bf{67.3}&	\bf{70.5}	&{71.3}	&77.8	&50.7	&60.4	&\bf{63.0}	&\bf{62.8} \\
				\hline
			\end{tabular}
		\end{center}
		\label{tab:S3DIS_6F}
	\end{table*}
	
	\begin{table*}[!htp]
		\renewcommand\tabcolsep{2.5pt} 
		\caption{Semantic segmentation results on the \textit{SemanticKITTI}.}
		\begin{center}
			\begin{tabular}{lc|ccccccccccccccccccc}
				\hline
				
				Method&
				\rotatebox{90}{mIoU}&
				\rotatebox{90}{road}&
				\rotatebox{90}{sidewalk}&
				\rotatebox{90}{parking}&
				\rotatebox{90}{other-ground }&
				\rotatebox{90}{building}&
				\rotatebox{90}{car}&
				\rotatebox{90}{truck}&
				\rotatebox{90}{bicycle}&
				\rotatebox{90}{motorcycle}&
				\rotatebox{90}{other-vehicle }&
				\rotatebox{90}{vegetation}&
				\rotatebox{90}{trunk}&
				\rotatebox{90}{terrain}&
				\rotatebox{90}{person}&
				\rotatebox{90}{bicyclist}&
				\rotatebox{90}{motorcyclist}&
				\rotatebox{90}{fence}&
				\rotatebox{90}{pole}&
				\rotatebox{90}{traffic sign} \\
				\hline
				\hline
				
				PointNet~\cite{pointnet} &14.6& 61.6& 35.7& 15.8& 1.4& 41.4 &46.3 &0.1 &1.3 &0.3 &0.8 &31.0 &4.6 &17.6 &0.2 &0.2 &0.0 &12.9 &2.4 &3.7\\
				SPGraph~\cite{SPGraph}& 17.4 &45.0 &28.5 &0.6 &0.6 &64.3 &49.3 &0.1 &0.2 &0.2 &0.8& 48.9 &27.2 &24.6& 0.3 &2.7 &0.1 &20.8& 15.9 &0.8\\
				SPLATNet~\cite{su2018splatnet}&18.4 &64.6& 39.1 &0.4 &0.0 &58.3 &58.2 &0.0 &0.0 &0.0& 0.0 &71.1& 9.9 &19.3 &0.0& 0.0& 0.0 &23.1 &5.6 &0.0\\
				PointNet++~\cite{pointnet2} &20.1& 72.0 &41.8 &18.7& 5.6 &62.3 &53.7 &0.9 &1.9 &0.2& 0.2& 46.5 &13.8 &30.0 &0.9 &1.0 &0.0 &16.9 &6.0 &8.9\\
				TangentConv~\cite{TangentConv}&40.9 &83.9 &63.9 &\bf{33.4} &\bf{15.4} &\bf{83.4} &\bf{90.8} &15.2&\bf{ 2.7}& 16.5 &12.1 &79.5 &49.3 &58.1 &23.0 &28.4 &\bf{8.1} &\bf{49.0} &35.8 &28.5\\
				\hline
				PointASNL & \bf{46.8}&\bf{87.4}&
				\bf{74.3}&
				24.3&
				1.8&
				83.1&
				87.9&
				\bf{39.0}&
				0.0&
				\bf{25.1}&
				\bf{29.2}&
				\bf{84.1}&
				\bf{52.2}&
				\bf{70.6}&
				\bf{34.2}&
				\bf{57.6}&
				0.0&
				\bf{43.9}&
				\bf{57.8}&
				\bf{36.9} \\
				\hline
			\end{tabular}
		\end{center}
		\label{tab:KITTI}
	\end{table*}

	\begin{figure*}[h]
		\begin{center}
			\includegraphics[width=1 \linewidth, height=8cm]{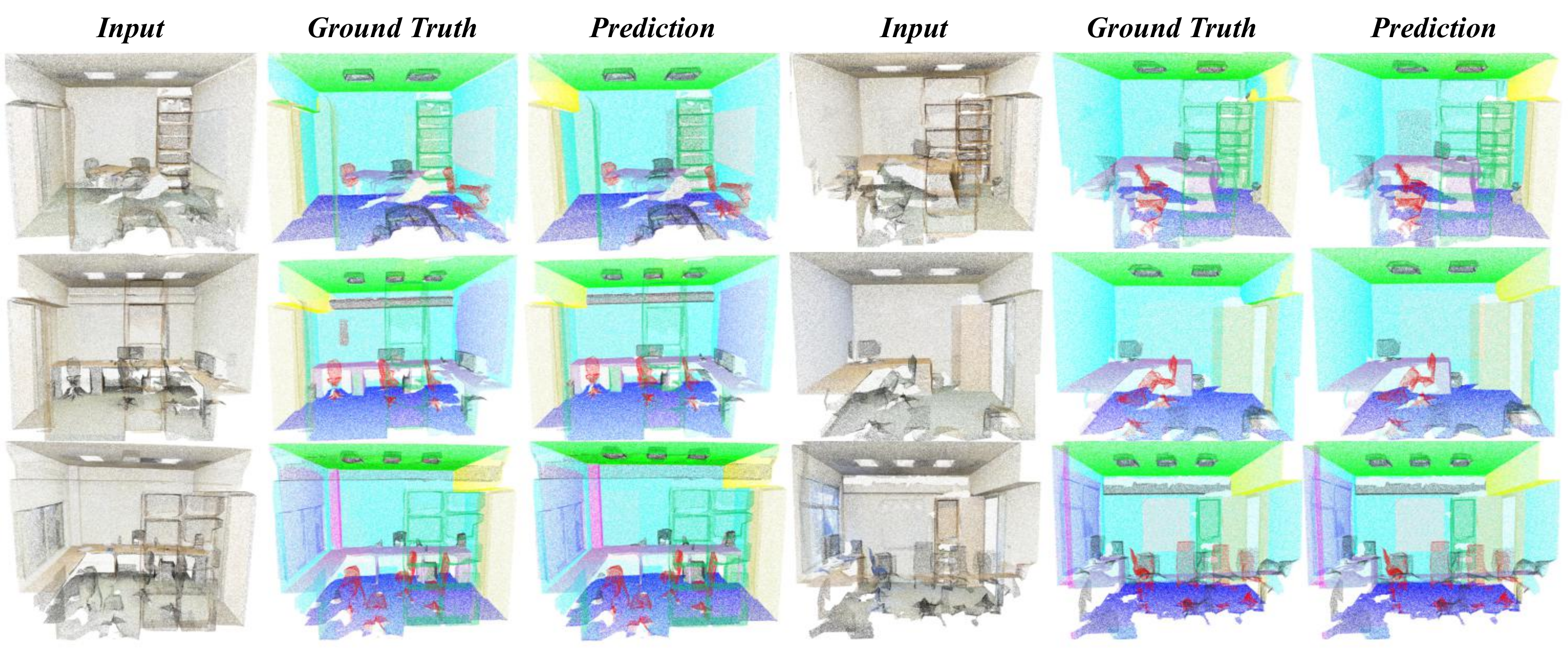}
		\end{center}
		\caption{More examples on \textit{S3DIS} datasets.}
		\label{fig:s3dis}
	\end{figure*}
	\begin{figure*}[h]
		\begin{center}
			\includegraphics[width=1 \linewidth, height=10cm]{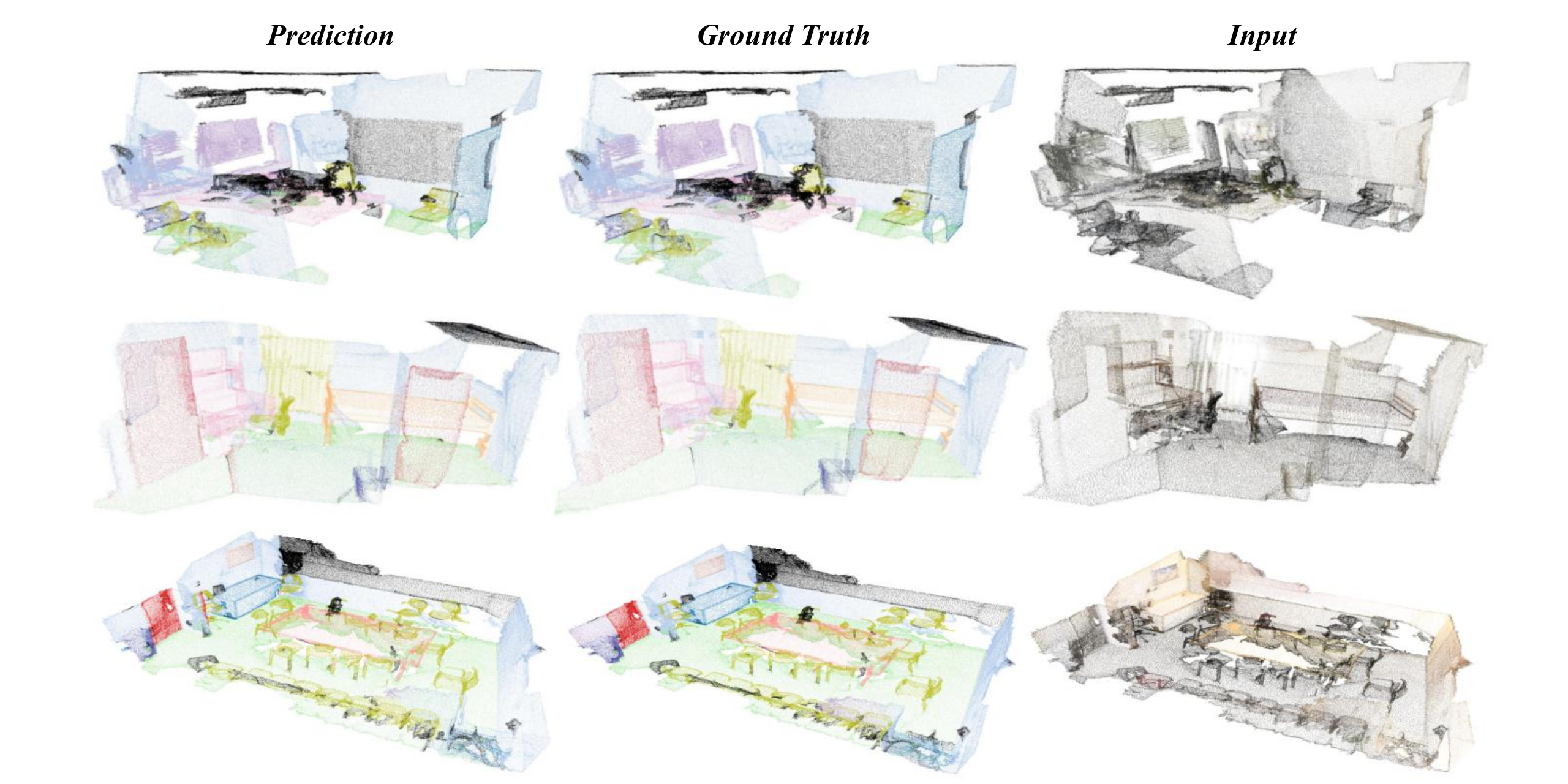}
		\end{center}
		\caption{More examples on \textit{ScanNet} datasets.}
		\label{fig:scannet}
	\end{figure*}
	\begin{figure*}[h]
		\begin{center}
			\includegraphics[width=1 \linewidth, height=10cm]{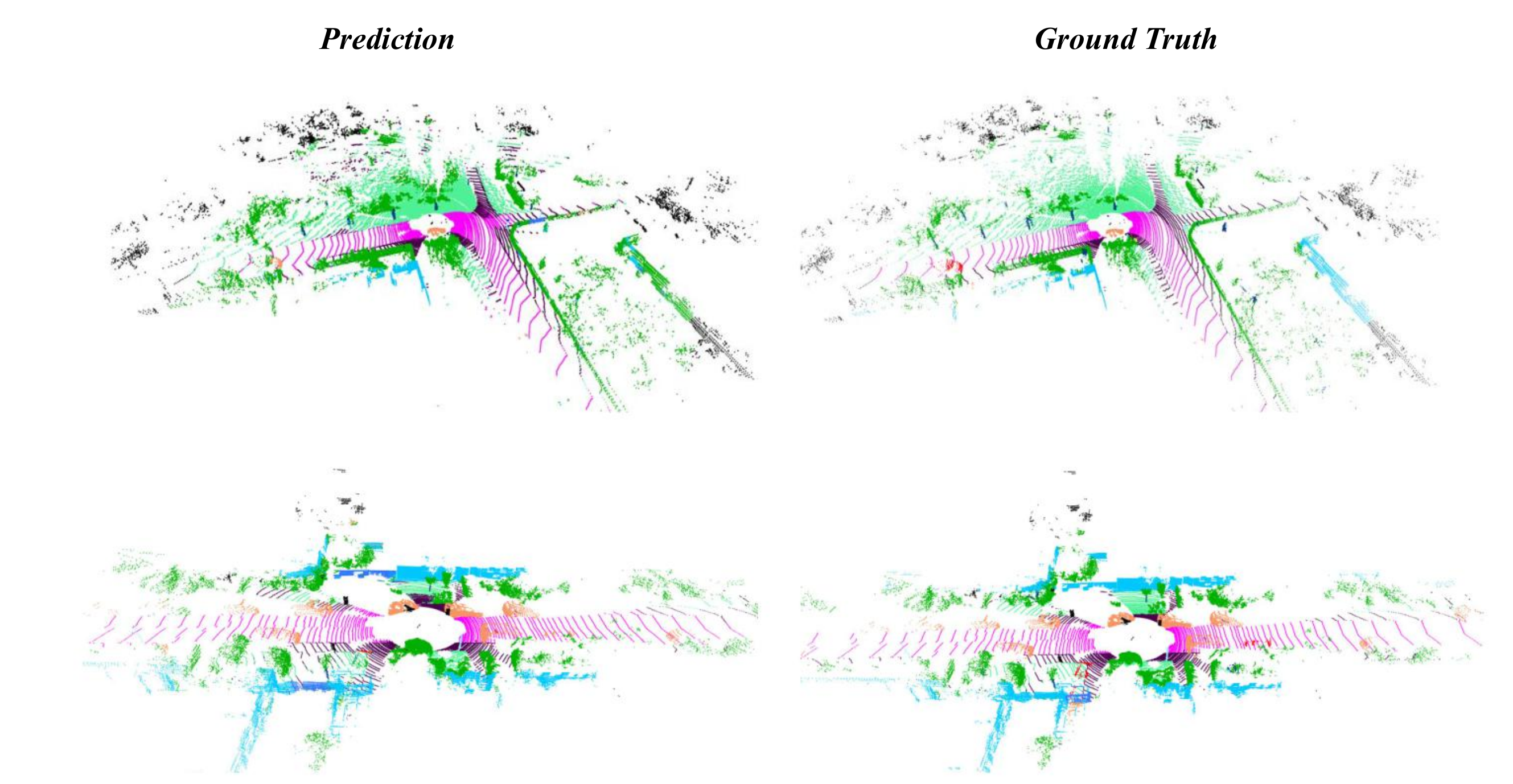}
		\end{center}
		\caption{More examples on \textit{SemanticKITTI} datasets.}
		\label{fig:KITTI}
	\end{figure*}
	\begin{figure*}[h]
		\begin{center}
			\includegraphics[width=1 \linewidth, height=10cm]{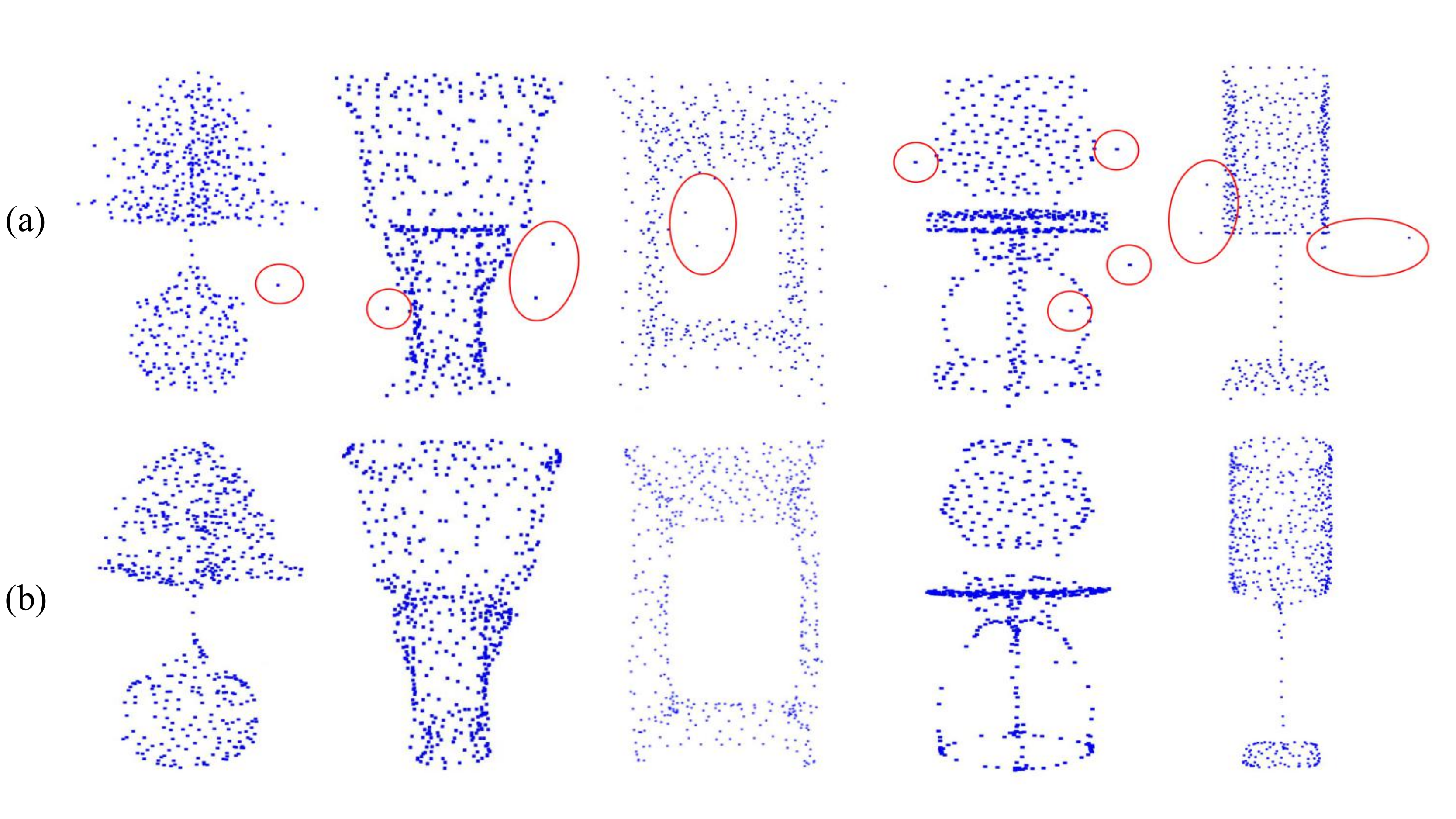}
		\end{center}
		\caption{Visualized results of AS module. (a) Sampled points via  farthest point sampling (FPS). (b) Sampled points ajusted by AS module.}
		\label{fig:as}
	\end{figure*}

\end{document}